\documentclass[10pt,twocolumn,letterpaper]{article}

\usepackage{cvpr}              
\definecolor{cvprblue}{rgb}{0.21,0.49,0.74}
\usepackage[pagebackref,breaklinks,colorlinks,allcolors=cvprblue]{hyperref}
\usepackage{rotating}
\usepackage{utfsym}
\usepackage{pifont}
\usepackage{colortbl}
\usepackage{url}
\usepackage{multirow}
\usepackage[accsupp]{axessibility}
\definecolor{First}{rgb}{0.95, 0.62, 0.61}
\definecolor{Second}{rgb}{0.97,0.81,0.63}
\definecolor{Third}{rgb}{1.0, 0.97, 0.70}

\def\paperID{38012} 
\def\confName{CVPR}
\def\confYear{2026}

\title{MetricHMSR:\\
Metric Human Mesh and Scene Recovery from Monocular Images}

\author{
Chentao Song$^{*1}$, He Zhang$^{*1}$, Haolei Yuan$^{3}$,  Haozhe Lin$^{1}$, Jianhua Tao$^{1}$, Hongwen Zhang$^{\dag2}$, Tao Yu$^{\dag1}$ \\
$^{1}$Tsinghua University, $^{2}$Beijing Normal University, $^{3}$Beihang University\\
}

\begin{document}
\twocolumn[{%
\renewcommand\twocolumn[1][!htb]{#1}%
\maketitle
\vspace{-9mm}
\begin{center}
    \includegraphics[width=0.9\linewidth]{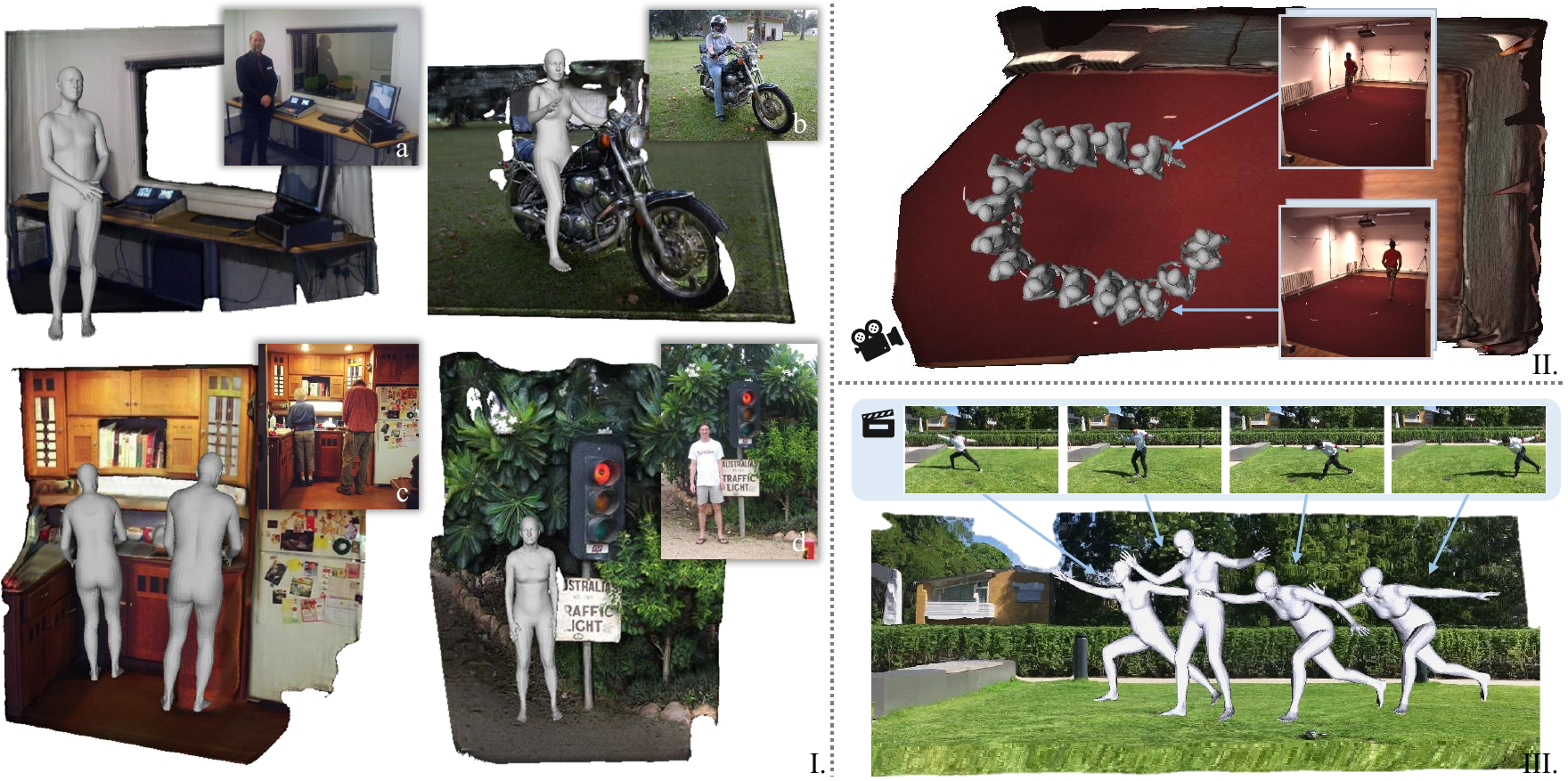}
    \vspace{-2mm}
    {\captionsetup{type=figure,hypcap=false}
    \caption{%
    \textbf{MetricHMSR} (\textbf{Metric} \textbf{H}uman \textbf{M}esh and \textbf{S}cene \textbf{R}ecovery) reconstructs human pose, metric shape, global position, as well as scene geometry from a monocular image. 
    Left: metric-consistent human–scene reconstruction. 
    Right: globally consistent 3D trajectories obtained by applying MetricHMSR independently to each frame.
    }

    \label{fig:teaser}}
    \vspace{-1mm}
\end{center}%
}]
\renewcommand{\thefootnote}{\fnsymbol{footnote}}
\footnotetext[1]{Equal contribution.}
\footnotetext[2]{Corresponding author.}
\renewcommand{\thefootnote}{\arabic{footnote}}
\begin{abstract}
We introduce MetricHMSR, a novel framework for recovering metric human meshes and 3D scenes from a single monocular image. 
Existing methods struggle to recover metric scale due to monocular scale ambiguity and weak-perspective camera assumptions. Moreover, their fully coupled feature representations make it difficult to disentangle local pose from global translation, often requiring multi-stage pipelines that introduce accumulated errors.
To address these challenges, we propose MetricHMR (\textbf{Metric} \textbf{H}uman \textbf{M}esh \textbf{R}ecovery), which incorporates a bounding camera ray map representation to provide explicit metric cues for human reconstruction, together with a Human Mixture-of-Experts (HumanMoE) that dynamically routes image features to specialized experts, enabling the disentangled perception of local human pose and global metric position.
Leveraging the recovered metric human as a geometric anchor, we further refine monocular metric depth estimation to achieve more accurate 3D alignment between humans and scenes. Comprehensive experiments demonstrate that our method achieves state-of-the-art performance on both human mesh recovery and metric human–scene reconstruction. Project Page: \href{https://Metaverse-AI-Lab-THU.github.io/MetricHMSR}{https://Metaverse-AI-Lab-THU.github.io/MetricHMSR}.

\end{abstract}    
\section{Introduction}
\label{sec:intro}
Human mesh~\cite{bogo2016smplify, kanazawa2018hmr, kolotouros2019spin, goel2023hmrv2, cai2024smpler} and scene recovery~\cite{Wang2025vggt,keetha2025mapanything,chen2025human3r} (HMSR) from a single image is a fundamental problem in computer vision and graphics. In particular, physically consistent metric reconstruction is crucial for emerging paradigms such as physical AI and embodied AI.

Since HMR~\cite{kanazawa2018hmr} has similarly pioneered the first end-to-end approach for 3D human pose and shape estimation.
The field has progressively advanced towards achieving higher precision in both 2D overlay~\cite{kolotouros2019spin,zhang2021pymaf, zhang2023pymafx} and local pose estimation~\cite{kocabas2020vibe,goel2023hmrv2,cai2024smpler,dwivedi2024tokenhmr,sarandi2025neural}. However, metric properties, such as metric shape and position, have rarely been considered. Bridging the gap from local to global and from plausible to metric remains challenging.

Early approaches adopt weak-perspective assumptions to simplify camera projection~\cite{kocabas2020vibe,goel2023hmrv2,kolotouros2019spin,zhang2021pymaf,zhang2023pymafx}. Subsequent works explore improved camera modeling~\cite{ye2023slamhmr,wang2023zolly,kocabas2021spec,patel2024camerahmr}, while others~\cite{wang2024tram,Baradel2024multihmr} assume full-perspective projection with known camera intrinsics when available. Notably, CLIFF~\cite{li2022cliff} is the first method to explicitly account for the influence of bounding boxes on both human rotation and global position. However, these approaches either employ camera models that deviate significantly from real-world imaging conditions or fail to capture the role of camera intrinsics in metric reconstruction and to disentangle features related to local pose and global position.
Another line of methods~\cite{ye2023slamhmr,sun2023trace,shen2024gvhmr,shin2024wham} attempts to introduce additional trajectory estimation modules to derive plausible human positions and motion trajectories. \cite{wang2024tram,wang2024blade} directly integrate monocular metric depth estimation (MMDE) approaches~\cite{bhat2023zoedepth,yang2024dav2} to obtain metrical results. However, introducing additional modules increases system complexity, while reliance on external monocular depth estimation inherently limits performance to that of the depth predictor.

Besides, the Vision Transformer (ViT) has garnered significant attention due to its exceptional performance\cite{dwivedi2024tokenhmr}. It divides the input image into distinct patches; however, not all regions contribute equally to metric human mesh recovery~\cite{zeng2022not}, as different regions carry distinct types of information. Current architectures process all tokens holistically, struggling to capture the nuanced similarities and differences between image patches.

To overcome the aforementioned problems, our insights are: (1) The camera intrinsic parameters and bounding box information are associated with the 3D position of the human. (2) Recent foundation models~\cite{Wang2025vggt} have validated that all key 3D attributes can be learned by a unified architecture without incorporating specialized modules. (3) We argue that feature disentanglement is beneficial.

Hence, we first propose MetricHMR, a module that simultaneously estimates local pose, metric shape, and metric position. Following~\cite{mildenhall2021nerf}, we represent camera parameters using pixel-aligned camera rays and construct a bounding ray map for the cropped image. The cropped image and the corresponding bounding ray map are jointly fed into the network, where the representation implicitly encodes both camera intrinsics and bounding box information, mitigating the location ambiguity introduced by image cropping~\cite{li2022cliff} and providing explicit metric cues for reconstruction.

To jointly model pose, metric shape, and metric position, we propose the HumanMoE. Built upon the MoE framework~\cite{muqeeth2023soft}, HumanMoE dynamically routes features to specialized experts, enabling feature-level disentanglement while preserving shared representations.
HumanMoE consists of a Patch MoE and a Global MoE, which capture complementary local and global context.
The Patch MoE processes image patches to learn specialized representations for different semantic regions, such as body parts or background elements, while the Global MoE captures holistic image-level information.
Together, they enable accurate recovery of metric human mesh from a single image.


Finally, metric human mesh provides a geometric anchor for scene understanding. We further introduce a human-guided metric depth refinement module that learns per-pixel depth transformations conditioned on the reconstructed human mesh.
This module refines MMDE and produces more physically consistent human–scene compositions.
By integrating metric human reconstruction with human-guided scene perception, we obtain a unified framework termed MetricHMSR, which enables metric-scale reconstruction of both humans and scenes from a single image.

In summary, our contributions are as follows:
\begin{itemize}
\item We propose MetricHMR, a novel human mesh recovery method that encodes metric cues via bounding camera rays and employs HumanMoE to disentangle features of local pose and global position features.

\item We propose MetricHMSR, a unified framework for metric human mesh and scene recovery. It further introduces a human-guided metric depth refinement module that leverages reconstructed human mesh priors to produce consistent metric depth and human–scene reconstruction.

\item Comprehensive experiments demonstrate that MetricHMSR achieves state-of-the-art (SOTA) performance in both human mesh recovery and MMDE. 
\end{itemize}
\section{Related Work}\label{sec:related}
\subsection{Human Mesh Recovery with Camera Modeling}
Human Mesh Recovery is a classic task that has seen significant progress. HMR~\cite{kanazawa2018hmr} is the first end-to-end method to regress 3D human pose and shape. Numerous methods have emerged, focusing on improving the accuracy of 2D overlay~\cite{kolotouros2019spin,zhang2021pymaf, zhang2023pymafx} and local pose estimation~\cite{Mueller2021tuch, kocabas2020vibe,goel2023hmrv2,cai2024smpler,dwivedi2024tokenhmr,sarandi2025neural}.
Most of the methods above primarily rely on the weak-perspective projection assumption. Researchers~\cite{kissos2020beyond,dwivedi2024tokenhmr} have gradually recognized that the incorrect camera model is a key limiting factor preventing further improvements in accuracy.

\begin{figure*}
  \centering
   \includegraphics[width=\linewidth]{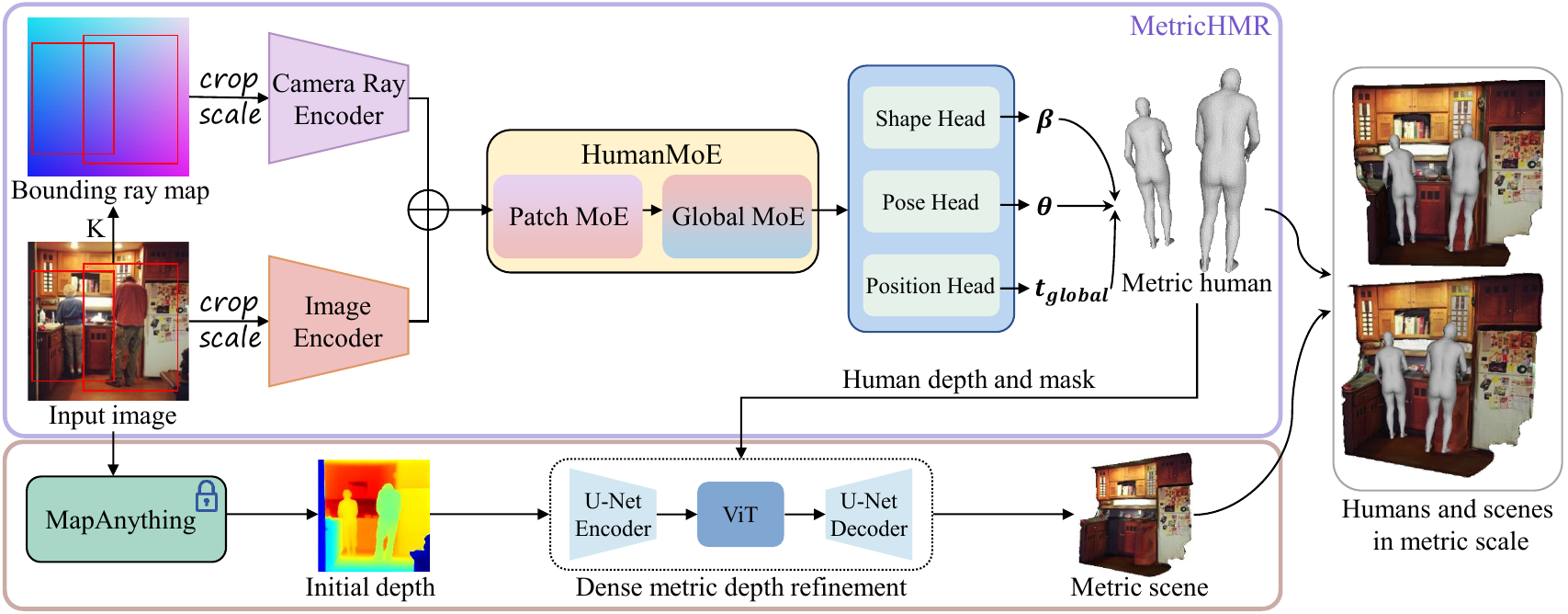}
   \caption{
   Overview of MetricHMSR. Given a single image, the framework jointly recovers the metric human mesh and the scene. The cropped image and the corresponding bounding ray map are encoded into tokens and processed by HumanMoE, which consists of a Patch MoE and a Global MoE to capture patch-level and image-level representations. The output heads predict the SMPL pose, shape, and global position. The recovered metric human mesh is then used to refine the depth predicted by MapAnything, producing geometrically consistent metric depth and enabling accurate human–scene reconstruction. $\oplus$ denotes concatenation.
   }
   \label{fig:4_pipeline}
\end{figure*}

To tackle this limitation, some methods~\cite{li2022cliff,wang2023zolly,kocabas2021spec,ye2023slamhmr,patel2024camerahmr,kissos2020beyond, wang2025prompthmr} have proposed various approaches for camera modeling or intrinsic parameter estimation, primarily by positing a relationship between the focal length and the dimensions of the full image. These methods enhance accuracy in scenarios with unknown camera parameters, albeit with a camera representation that lacks precision. This has prompted another branch of research~\cite{li2022cliff, Baradel2024multihmr, wang2024tram,hao2025perspose} to directly employ full perspective projection and ground-truth camera intrinsics, wherever possible. By more closely approximating the real camera, this model minimizes projective errors~\cite{dwivedi2024tokenhmr} and holds significant potential, especially given that many methods~\cite{piccinelli2024unidepth,Wang2025vggt,jang2025pow3r} can now estimate intrinsic parameters from images.

\subsection{Human Mesh Recovery in Global Space}
Beyond estimating the human body's own pose, reconstructing its global position and trajectory in space is also essential.
Methods relying solely on vision usually incorporate a dedicated global estimation module, distinct from camera pose estimation, specifically for estimating 3D global position and trajectory.

Sun et al.~\cite{sun2022bve, sun2023trace} introduce the Bird’s Eye View to infer 3D position. SLAMHMR~\cite{ye2023slamhmr} uses DROID-SLAM~\cite{teed2021droid} to estimate camera pose and introduces an optimization module to obtain 3D motion trajectory. WHAM~\cite{shin2024wham} incorporates the trajectory decoder and refiner to regress the global trajectory. GVHMR~\cite{shen2024gvhmr} proposes the Gravity-View coordinate system to estimate the global motion.

The development of monocular metric depth estimation~\cite{bhat2023zoedepth,yin2023metric3d, piccinelli2024unidepth,yang2024dav2} has brought new opportunities and inspirations to human mesh recovery.
 BLADE~\cite{wang2024blade} utilizes DepthAnythingV2~\cite{yang2024dav2} to estimate the depth of the pelvis near the camera, to accurately recover the focal length and 3D translation.
 TRAM~\cite{wang2024tram} uses ZeoDepth~\cite{bhat2023zoedepth} to estimate the metric scale of humans. 
 Human3R~\cite{chen2025human3r}, built upon CUT3R~\cite{wang2025cut3r}, a 4D metric reconstruction method, achieves 4D human–scene reconstruction in the world frame.
Methods relying on existing modules are inherently constrained by their performance. We empirically observe that recovering a metric human mesh does not require building upon or introducing additional specialized modules.

Our method, MetricHMSR, can recover the local pose, metric shape, and metric position with a unified model without introducing specialized modules.

\subsection{Metric Scene Recovery from Monocular Image}
Monocular metric depth estimation~\cite{bhat2023zoedepth, yin2023metric3d, zhao2024metric, goel2023hmrv2, piccinelli2024unidepth, pham2024sharpdepth, yang2024dav2} has long been a fundamental challenge in computer vision. By leveraging camera intrinsic parameters, metric scene reconstruction can be achieved.
The development of metric depth estimation has brought new opportunities and inspirations to human pose estimation. 
Recent advances in large feed-forward transformer models, such as VGGT~\cite{Wang2025vggt}, demonstrate the capability to directly infer all key 3D attributes of scenes from single images. 
Human3R~\cite{chen2025human3r} can reconstruct both 3D scenes and human bodies. MapAnything \cite{keetha2025mapanything} regresses the metric 3D scene geometry and cameras, achieving the current state-of-the-art performance.

Based on MapAnything, we utilized the human mesh as a guide to achieve more accurate metric scene reconstruction.
This ultimately achieves the recovery of overlayed human and 3D scene in metric scale.
\section{MetricHMSR}\label{sec:method}
We proposed MetricHMSR to regress the metric human mesh and scene from a monocular image. MetricHMSR consists of two components: MetricHMR for metric human mesh recovery, and a human-guided metric depth refinement module for scene reconstruction. The pipeline is shown in \cref{fig:4_pipeline}.

\subsection{MetricHMR}\label{sec:metricHMR}
We use SMPL~\cite{loper2015smpl}, a linear parametric model, to represent the human body. Given a single RGB image $I$ as input, our goal is to recover the SMPL pose parameters $\theta \in \mathbb{R}^{72}$, shape parameters $\beta \in \mathbb{R}^{10}$, and translation $t_{global} \in \mathbb{R}^{3}$, corresponding to the local pose, metric body shape, and metric 3D position of the human body.


To this end, we propose MetricHMR, a MoE-based framework for metric human mesh recovery, as illustrated in \cref{fig:4_pipeline}. MetricHMR consists of two key components: a bounding ray map that provides explicit metric cues for 3D position, and HumanMoE that enables feature-level disentanglement between local pose and global position.
The network takes the cropped image and the corresponding bounding camera ray map as input. The ray map is constructed from the camera intrinsic parameters following the formulation described in \cref{sec:ray}. When intrinsics are unavailable, they are estimated using AnyCalib~\cite{tirado2025anycalib}.

For feature extraction, we employ ViTPose~\cite{xu2022vitpose} and ViT-Large-Patch16-224~\cite{dosovitskiy2021vit} as the respective encoders for the image and the ray map. The encoded features are concatenated and jointly passed through the proposed HumanMoE modules to capture both local and global feature representations. Finally, MLP heads regress the pose $\theta$, shape $\beta$, and metric 3D translation $t_{global}$.
\subsection{Bounding Ray Map}\label{sec:ray}
Existing human mesh recovery methods suffer from the following limitations: (1) Focal length influences the perception of metric distance, and the subject's 2D image position correlates with global 3D location (see Supp. Mat. for details). However, current networks struggle to effectively leverage this information for metric human mesh recovery. (2) Cropping and scaling operations impair the network's ability to perceive metric information. 

\begin{figure}[h]
  \centering
   \includegraphics[width=\linewidth]{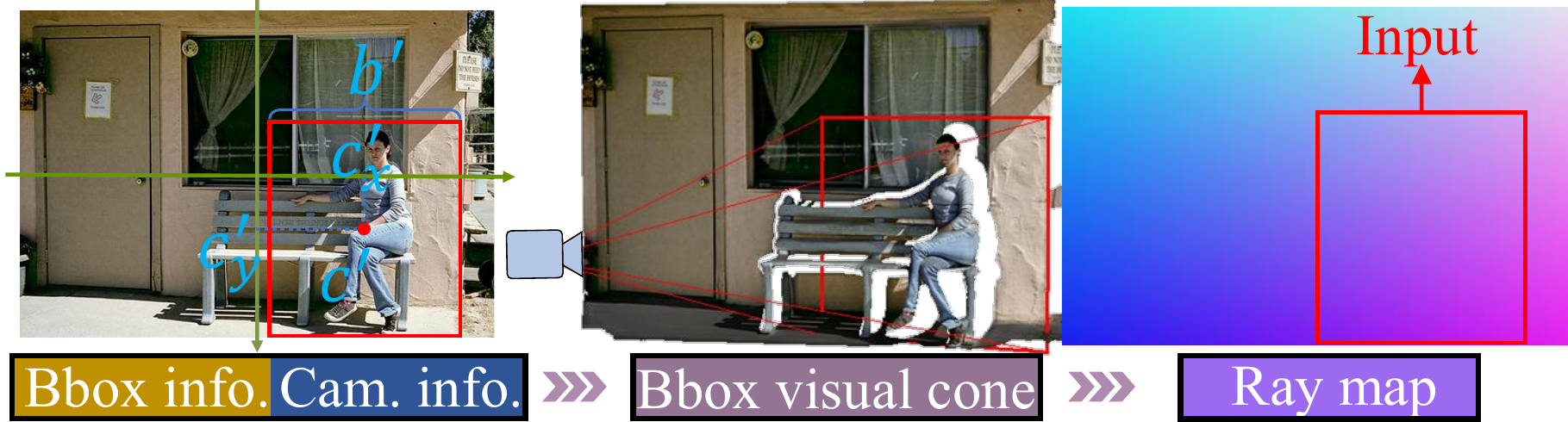}
   \caption{Bounding ray map representation. Left: bounding box and camera representation in CLIFF. Middle: bounding box visual cone. Right: camera ray representation. }
   \label{fig:4_camera_ray}
\end{figure}

To address the aforementioned problems, we incorporate the bounding ray map, a pixel-aligned camera representation, into our framework to comprehensively represent human image position, camera intrinsics, and the effects of cropping and scaling operations, as shown in~\cref{fig:4_camera_ray}.

Assuming the intrinsic parameters of the original image is $K$, with focal lengths $(f_x, f_y)$ and principal point $(c_x, c_y)$.
Following~\cite{mildenhall2021nerf}, the camera ray $d$ corresponding to a pixel $(u,v)$ on the image is calculated as:
\begin{equation}
  d = K^{-1}[u,v,1].
  \label{eq:ray}
\end{equation}

Assuming the top-left corner of the cropping bounding box is at $(u_{bbox},v_{bbox})$ and the scaling factor is $s$, the intrinsic matrix $K^{\prime}$ of the transformed image can be expressed as:
\begin{equation}
K^{\prime} =
\left[
\begin{array}{ccc}
    f_x/s & 0 & (c_x-u_{bbox})/s \\
    0 & f_y/s & (c_y-v_{bbox})/s \\
    0 & 0 & 1
\end{array}
\right].
\label{eq:crop_k}
\end{equation}
Then, we use Eq.\eqref{eq:ray} to calculate the camera ray bundle corresponding to the transformed image.
For images with unknown camera parameters, we approximate the focal length using the longer side of the image and set the principal point to the image center.

\subsection{HumanMoE}
Existing human mesh recovery methods commonly employ dense multi-layer perceptrons (MLP)~\cite{kanazawa2018hmr,li2022cliff} or transformers~\cite{wang2024tram} as their decoder. However, these approaches may present two potential problems: (1) a single decoder feature may be insufficient to represent the variations of local pose, metric shape, and metric 3D position across diverse scenarios; and (2) dense networks may struggle to decouple hierarchical image features.
Given the demonstrated power of MoE in foundation models~\cite{dai2024deepseekmoe,liu2024deepseekv2}, 
we propose HumanMoE, a unified module to comprehensively learn human-related knowledge across different dimensions.

Mixture-of-Experts (MoE) is commonly composed of two core components: (1) MoE Layer: Consists of $N$ feed-forward networks (FFNs), each referred to as an "expert" ($e_i(\cdot), i=1,2,...,N$). These experts operate as independent modules, each designed to learn specialized domain knowledge.
(2) Router: Functions as a gating module, implemented by a learnable MLP. 
Given the token $x$, it automatically computes the probability $g_i(x)$ of it being assigned to each expert.
The final output of the MoE is derived from a linear weighted combination of the outputs of the selected experts and their corresponding router weights. The output of the corresponding top-$K$ experts is
\begin{equation}
  \mathrm{MoE}(x)=\sum_{i=0}^K g_i(x)e_i(x),
  \label{eq:moe}
\end{equation}

\begin{figure}
  \centering
   \includegraphics[width=0.72\linewidth]{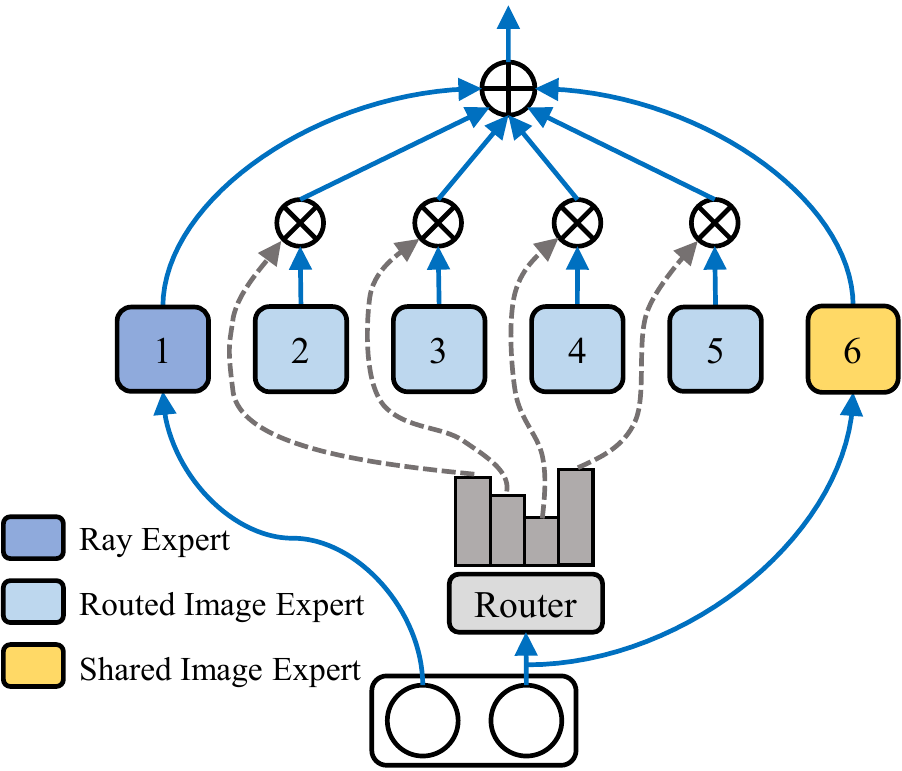}
   \caption{The architecture of MoE Layer. We designed a ray expert to learn features from camera rays, 4 routed image experts to process specialized image knowledge, and a shared image expert to capture common image knowledge.}
   \label{fig:4_moe}
\end{figure}

\paragraph{MoE layer design.} Here we employ the soft MoE~\cite{muqeeth2023soft} to replace the feed-forward networks (FFNs) in dense transformer blocks.
Compared to standard MoE, where each token is only processed by the top-$K$ experts, Soft MoE assigns a weighted combination of tokens to each expert. This approach demonstrates better training stability and expert scalability. We also introduce shared experts~\cite{dai2024deepseekmoe,han2024vimoe} that remain active for all tokens and are designed to capture fundamental common knowledge. Shared expert prevents redundant learning of generic patterns across specialized experts, thereby optimizing model capacity utilization.
We design 4 routed experts to learn specialized visual knowledge, and a shared expert to capture common visual knowledge. We also design a ray expert specifically for processing ray tokens.
The architecture is illustrated in \cref{fig:4_moe}.

\paragraph{HumanMoE.} HumanMoE comprises two components: a Patch MoE and a Global MoE, which model representations at different levels. In metric human mesh recovery, image patches play distinct roles and vary in importance. To capture this heterogeneity, we introduce the Patch MoE, which dynamically routes patches to specialized experts based on semantic content. Patches with similar semantics or roles are processed by the same expert, enabling effective and explicit feature-level disentanglement. 
Building on this, we further introduce the Global MoE to aggregate full-image representations and capture global contextual information. This complements the Patch MoE by enabling consistent reasoning about metric properties across the entire image.

\begin{figure}
  \centering
   \includegraphics[width=\linewidth]{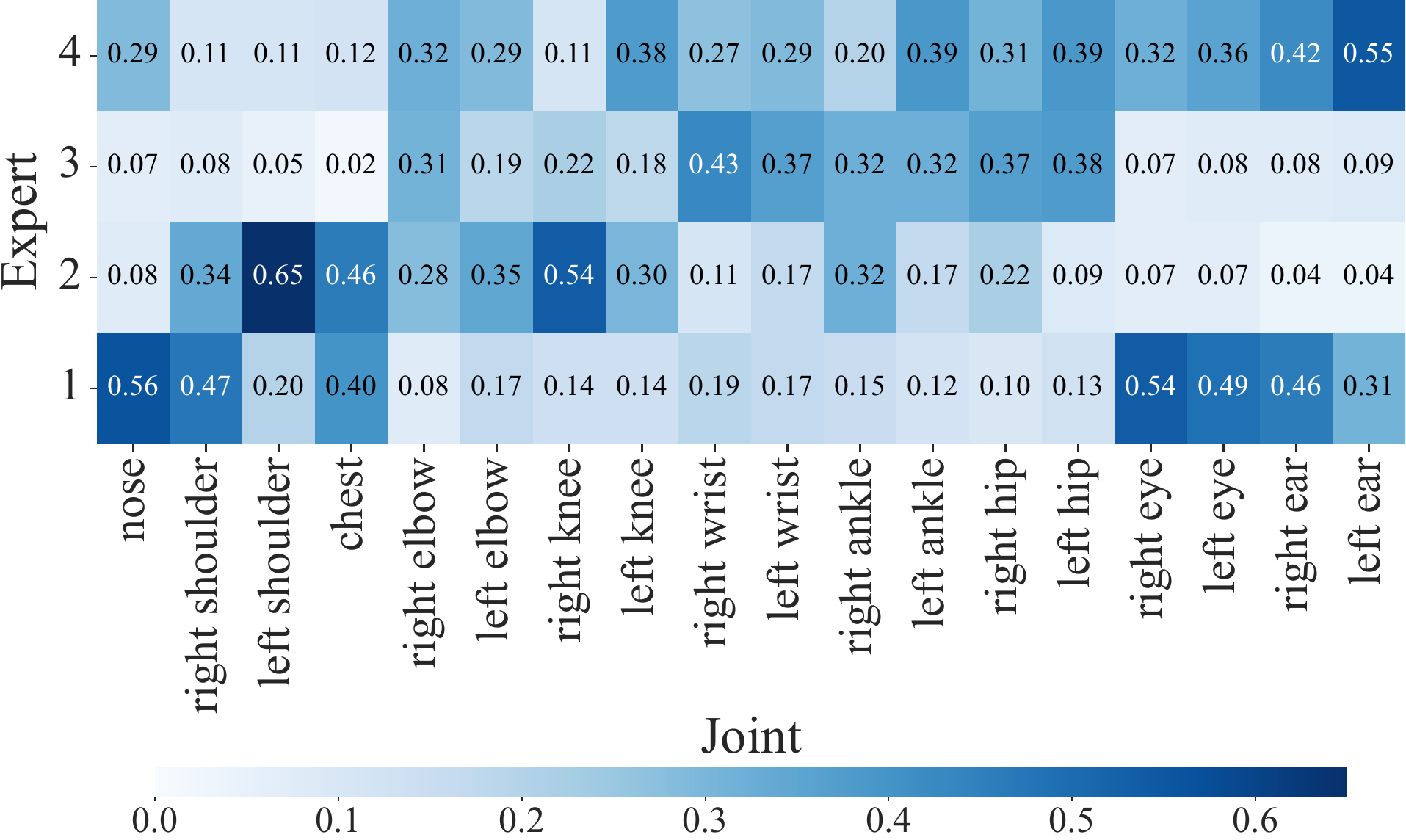}
   \caption{Routing heatmap of the deepest (last) MoE layer for image feature on 3DPW.}
   \label{fig:5_expert_heatmap}
\end{figure}

\paragraph{Token routing.} 
Here, we analyze the routing behavior of HumanMoE and its role in disentangling features related to local pose and global position.
The deepest (last) MoE layer exhibits the strongest semantic representation~\cite{han2024vimoe}. 
For Patch MoE, distinct body regions are consistently assigned to different experts, while similar body parts across images are predominantly routed to the same experts, indicating emergent semantic specialization.
To quantify this behavior, we analyze the expert assignments of patches containing 2D human keypoints on the 3DPW dataset and compute the corresponding routing statistics, as shown in \cref{fig:5_expert_heatmap}. Despite potential noise from keypoint detection errors and coarse patch partitioning, the large-scale statistics provide reliable evidence for this observation.
For Global MoE, we analyze details in the supplementary materials.


\paragraph{MoE Loss.}
Following prior works~\cite{shazeer2017outrageously,fedus2022switch}, to encourage balanced expert utilization, we add a soft load-balancing loss on the gating probabilities. Let $p_i$ denote the average routing probability of expert $i$ over all tokens in a batch. The auxiliary loss is defined as
\begin{equation}
\mathcal{L}_{\text{aux}} = \lambda\, K \sum_{i=1}^{K} p_i^{2},
\end{equation}
where $K$ is the number of experts and $\lambda$ is the weights. Since $\sum_i p_i = 1$, this term is minimized when all experts are used uniformly, i.e., $p_i = 1/K$, and increases when routing collapses onto a few experts, thus promoting a more even load distribution. This loss is introduced solely to balance the usage among the routed image experts.

\paragraph{Over-complete Loss Design for MetricHMR.}
 We train our model with the following losses:
\begin{align}
    \begin{aligned}
          \mathcal{L} =& \lambda_{J_{2D}}\mathcal{L}_{J_{2D}}+\lambda_{J_{3D}}\mathcal{L}_{J_{3D}}+\lambda_{V_{3D}}\mathcal{L}_{V_{3D}}+\\
&\lambda_{\theta}\mathcal{L}_{\theta}+\lambda_{\beta}\mathcal{L}_{\beta}+\lambda_{h}\mathcal{L}_{h}.
    \end{aligned}
    \label{eq:hmr_losses}
\end{align}
$\mathcal{L}_{J_{2D}}$, $\lambda_{J_{3D}}$, $\mathcal{L}_{V_{3D}}$, $\mathcal{L}_{\theta}$, $\mathcal{L}_{\beta}$, and $\mathcal{L}_{h}$
are the loss of 2D keypoints, 3D joints, vertices, SMPL pose, SMPL shape and SMPL body height. $\lambda_{J_{2D}}$, $\lambda_{J_{3D}}$, $\lambda_{V_{3D}}$, $\lambda_{\theta}$, $\lambda_{\beta}$ and $\lambda_{h}$ are their corresponding weights. For detailed definitions of each loss term, please refer to the supplementary material.



In particular, VGGT~\cite{Wang2025vggt} suggests that over-complete predictions during training can bring substantial performance gains, even when some variables are related through closed-form relationships. Inspired by this idea, we adopt over-complete losses in our framework. For example, in human body modeling, height and shape parameters are inherently coupled. Prior work~\cite{choutas2022accurate} shows that height supervision improves the accuracy of human mesh regression.

\subsection{Human-Guided Metric Depth Refinement}
Human scale and body structure provide strong priors for MMDE.
Human~\cite{zhao2024metric} leverages this idea by inpainting humans into images, estimating human meshes with HMR 2.0~\cite{goel2023hmrv2}, and using them as metric references to scale monocular depth predictions.
However, since HMR 2.0 cannot recover metric body shape or global 3D position, and Metric from Human only achieves global scaling, its performance remains limited in accuracy.

In contrast, MetricHMR recovers human meshes in metric 3D space. This enables us to use the reconstructed metric human mesh as a geometric reference to refine MMDE. 
Consequently, the human mesh can align not only with the 2D image but also with the 3D scene geometry.

\begin{figure}[htbp]
  \centering
   \includegraphics[width=0.68\linewidth]{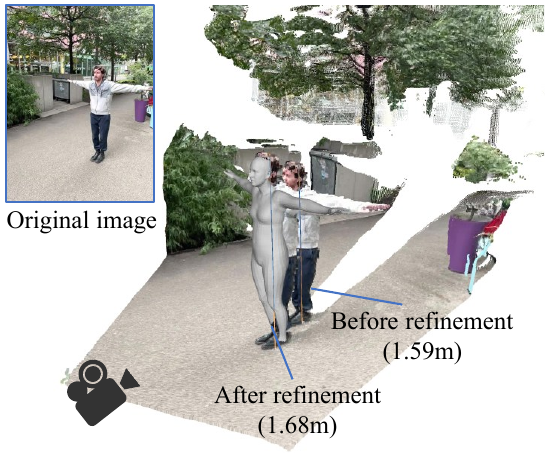}
   \caption{Illustration of human-guided metric depth refinement.}
   \label{fig:depth_refine}
\end{figure}

To this end, we introduce a per-pixel metric depth refinement module that calibrates monocular depth using metric human meshes as geometric references (Fig.~\ref{fig:depth_refine}).

Given a depth map \(z_{\text{in}}(x)\) predicted by MapAnything~\cite{keetha2025mapanything}, 
a hybrid UNet--ViT backbone predicts spatially varying affine fields \((s(x),\,b(x))\) to correct the depth:
\begin{equation}
\hat z(x)=s(x)\, z_{\text{in}}(x) + b(x).
\end{equation}
The predicted fields are locally adaptive while regularized to remain globally smooth and coherent, enabling metric-scale calibration with gentle local adjustments.

To anchor the solution to an absolute metric scale, we project the human mesh onto the image and extract the visible-surface depth per pixel to form a sparse anchor map \(z_{\text{hmr}}(x)\) with mask \(M_a(x)\).
An anchor consistency loss is applied at these pixels during training.
The overall loss is
\begin{equation}
\mathcal{L}
= \lambda_d\,\mathcal{L}_{\text{depth}}
+ \lambda_a\,\mathcal{L}_{\text{anchor}}
+ \lambda_{\mathrm{tv}}\,\mathcal{L}_{\mathrm{tv}}
+ \lambda_{\mathrm{var}}\,\mathcal{L}_{\mathrm{var}} \, ,
\end{equation}
where \(\mathcal{L}_{\text{depth}},\,\mathcal{L}_{\text{anchor}},\,\mathcal{L}_{\text{tv}},\,\mathcal{L}_{\text{var}}\)
denote the depth regression loss on valid pixels, the anchor consistency loss on HMR front-surface points,
the total-variation regularization on \((s,b)\), and the spatial variance regularization on \((s,b)\) to keep them near their global means, respectively. The \(\lambda\)'s are scalar weights.


\section{Experiments}\label{sec:exp}

\subsection{Implementation Details}

We train MetricHMR on BEDLAM~\cite{black2023bedlam}, AIC~\cite{wu2017ai}, COCO~\cite{lin2014coco}, MPII~\cite{andriluka2014mpii}, and 3DPW~\cite{von20183dpw} for 40 epochs with a weight decay of $1\times10^{-4}$. We use the AdamW optimizer with a batch size of 64 on a single NVIDIA A100 GPU, and set the learning rate to $1\times10^{-5}$. For the depth refinement network, we further train the model on the PROX RGB-D dataset ~\cite{hassan2019prox} to learn human-aware metric depth correction.

\subsection{Comparisons}
We present comprehensive evaluations of MetricHMSR, benchmarking it against state-of-the-art methods in human pose estimation. Following the established convention in Human3R~\cite{chen2025human3r}, we categorize the baselines into online and offline paradigms. Offline methods typically employ batch processing of complete video sequences to achieve globally consistent reconstructions, thereby pursuing globally optimal solutions. In contrast, online methods operate via a streaming framework that recursively reconstructs human motion as new frames arrive, maintaining low memory and computational footprints. Within this framework, we systematically evaluate the accuracy of our method in local pose estimation and global trajectory recovery. We further compare MetricHMSR against state-of-the-art methods under varying camera intrinsic settings. 


We first evaluate human mesh recovery from both local pose and global motion. 
For local pose estimation, we report per-vertex error (PVE), mean per-joint position error (MPJPE), and Procrustes-aligned MPJPE (PA-MPJPE). 
For global motion and trajectory estimation, we adopt WA-MPJPE$_{100}$ (WA-M.), W-MPJPE$_{100}$ (W-M.), Root Translation Error (RTE), and Egocentric-frame Root Velocity Error (ERVE). 
RTE is reported in percentage (\%), ERVE in mm/frame, and all other metrics are reported in mm.

\paragraph{Global trajectory estimation.} 
For global motion and trajectory estimation, MetricHMSR exhibits robust and competitive performance across both the EMDB-2~\cite{kaufmann2023emdb} (with dynamic camera) and RICH (with static camera) datasets under varying camera parameter conditions, as shown in \cref{tab:pred_extr}, \cref{tab:gt_extr}, and \cref{tab:static_extr}. Notably, in dynamic camera scenarios, MetricHMSR achieves performance on par with SOTA offline methods, regardless of whether camera parameters are estimated or ideal. Furthermore, on the RICH dataset, our method also exhibits highly competitive results. It achieves the best Root Translation Error (RTE) among all online and offline approaches, while also attaining strong performance in both WA-MPJPE and W-MPJPE metrics.

\begin{table}[htbp]
  \centering
  \setlength{\tabcolsep}{3.05pt} 
    \begin{tabular}{cccccc}
    \toprule
          &       & \multicolumn{4}{c}{EMDB-2} \\
\cmidrule{3-6}          &   Method    & WA-M. & W-M. & RTE   & ERVE \\
    \midrule
    \multirow{4}[1]{*}{\begin{sideways}offline\end{sideways}} 
        & SLAHMR~\cite{ye2023slamhmr}  & 326.9  &  776.1 & 10.2  & - \\
    & GVHMR~\cite{shen2024gvhmr} & 111.0   & 276.5 & 2.0     & - \\
          & TRAM~\cite{wang2024tram}  & \cellcolor{Third}76.4  & \cellcolor{Third}222.4 & \cellcolor{Second}1.4   & \cellcolor{First}10.3 \\
          & PromptHMR-vid~\cite{wang2025prompthmr} & \cellcolor{First}71.0    & \cellcolor{Second}216.5 & \cellcolor{First}1.3   & - \\
          \midrule
    \multirow{4}[1]{*}{\begin{sideways}online\end{sideways}} & TRACE~\cite{sun2023trace} & 529.0   & 1702.3 & 17.7  & 370.7 \\
          & WHAM~\cite{shin2024wham}  & 133.3 & 343.9 & 4.6   & \cellcolor{Third}14.7 \\
          & Human3R~\cite{chen2025human3r} & 112.2 & 267.9 & 2.2   & - \\
          & MetricHMSR & \cellcolor{Second}72.1  & \cellcolor{First}199.5 & \cellcolor{Second}1.4   & \cellcolor{Second}10.6 \\
    \bottomrule
    \end{tabular}%
      \caption{Quantitative comparisons of global motion and trajectory estimation with predicted extrinsic parameters on EMDB-2, a dynamic camera dataset. }
  \label{tab:pred_extr}
\end{table}%

\begin{table}[htbp]
  \centering
  \setlength{\tabcolsep}{3.05pt} 
    \begin{tabular}{cccccc}
    \toprule
          &       & \multicolumn{4}{c}{EMDB-2} \\
\cmidrule{3-6}          &   Method    & WA-M. & W-M. & RTE   & ERVE \\
    \midrule
    \multirow{4}[1]{*}{\begin{sideways}offline\end{sideways}} & GVHMR~\cite{shen2024gvhmr} & 109.1   & 274.9 & 1.9     & - \\
          & GLAMR~\cite{yuan2022glamr}  & 100.5  & 308.0 &  0.8  & 11.9 \\
          & TRAM~\cite{wang2024tram}  & \cellcolor{Third}62.7  & \cellcolor{Third}182.8 &  \cellcolor{First}0.2  & \cellcolor{Third}9.6  \\
          & PromptHMR-vid~\cite{wang2025prompthmr} & \cellcolor{First}52.9    & \cellcolor{First}142.1 & \cellcolor{First}0.2   & \cellcolor{First}7.8 \\
          \midrule
    \multirow{3}[1]{*}{\begin{sideways}online\end{sideways}} & TRACE~\cite{sun2023trace} & 131.4   & 422.0 & 1.4  & 18.8 \\
          & WHAM~\cite{shin2024wham}  & 131.1 & 335.3 & 4.1   & 13.7 \\
          & MetricHMSR & \cellcolor{Second}55.6  & \cellcolor{Second}152.5 & \cellcolor{First}0.2   & \cellcolor{Second}9.4 \\
    \bottomrule
    \end{tabular}%
      \caption{Quantitative comparisons of global motion and trajectory estimation with GT extrinsic parameters on EMDB-2, a dynamic camera dataset.}
  \label{tab:gt_extr}
\end{table}%

\begin{table}[htbp]
  \centering
  \setlength{\tabcolsep}{4pt} 
    \begin{tabular}{cccccc}
    \toprule
          &       & \multicolumn{3}{c}{RICH} \\
\cmidrule{3-5}          &    Method   & WA-M. & W-M. & RTE\\
    \midrule
    \multirow{6}[1]{*}{\begin{sideways}offline\end{sideways}}
    & COIN~\cite{li2024coin}  & 169.5  & 254.5 &  -   \\
    & GVHMR~\cite{shen2024gvhmr} & \cellcolor{First}78.8   & \cellcolor{First}126.3 & \cellcolor{Second}2.4      \\
      & TRAM~\cite{wang2024tram}  & 127.8  & 238.0 &  6.0  \\
      & GLAMR~\cite{yuan2022glamr}  & 129.4  & 236.2 &  3.8 \\
      & SLAHMR~\cite{ye2023slamhmr}  & 132.2  & 237.1 &  6.4  \\
      & PromptHMR-vid~\cite{wang2025prompthmr} & 118.2    & 207.0 & 5.6 \\
          \midrule
    \multirow{4}[1]{*}{\begin{sideways}online\end{sideways}} & TRACE~\cite{sun2023trace} & 238.1   & 925.4 & 610.4  \\
          & WHAM~\cite{shin2024wham}  & \cellcolor{Second}108.4 & 196.1 & 4.5 \\
          & Human3R~\cite{chen2025human3r}  & 110.0 & \cellcolor{Third}184.9 & \cellcolor{Third}3.3 \\
          & MetricHMSR &  \cellcolor{Third}109.6 & \cellcolor{Second}165.5 & \cellcolor{First}1.9   \\
    \bottomrule
    \end{tabular}%
      \caption{Quantitative comparisons of global motion and trajectory estimation on RICH, a static camera dataset.}
  \label{tab:static_extr}
\end{table}%

\paragraph{Local pose estimation.}  For human mesh recovery (local pose) evaluation, as summarized in \cref{tab:localpose}, our proposed MetricHMSR demonstrates highly competitive performance across both the 3DPW and EMDB-1~\cite{kaufmann2023emdb}. On 3DPW, MetricHMSR achieves the best results across all three metrics. On EMDB-1, MetricHMSR also delivers strong performance, ranking second in PA-MPJPE and PVE and third in MPJPE, remaining competitive and closely comparable to the top-performing method, PromptHMR.

\begin{table}[htbp]
\centering
\setlength{\tabcolsep}{2.1pt} 
\begin{tabular}{lcccccc}
    \toprule
    & \multicolumn{3}{c}{3DPW} & \multicolumn{3}{c}{EMDB-1}\\
    \cmidrule{2-7}     Method     & PA-M.  & M. & PVE & PA-M.   & M. & PVE\\
    \midrule
    HMR2.0a~\cite{goel2023hmrv2} & 44.4 & 69.8&	 82.2&	61.5&	 97.8&	120.0\\
    WHAM~\cite{shin2024wham}&	35.9&	57.8&	68.7&	50.4&	79.7&	94.4\\
    TRAM~\cite{wang2024tram}&	35.6&	59.3&	69.6&	45.7&	74.4&	86.6\\
    PersPose~\cite{hao2025perspose}&	39.1&	60.1&	72.4&	-&	-&	-\\
    Human3R~\cite{chen2025human3r}&	44.1&	72.1&	84.9&	48.5&	73.9&	86\\
    CameraHMR~\cite{patel2024camerahmr}&	\cellcolor{Second}35.1&	\cellcolor{Second}56.0&	\cellcolor{Second}65.9&	\cellcolor{Third}43.3&	\cellcolor{Second}70.3&	\cellcolor{Third}81.7\\
    PromptHMR~\cite{wang2025prompthmr}&	\cellcolor{Third}35.5&	\cellcolor{Third}56.9&	\cellcolor{Third}67.3&	\cellcolor{First}40.1&	\cellcolor{First}68.1&	\cellcolor{First}79.2\\
    MetricHMSR&	\cellcolor{First}33.6&	\cellcolor{First}53.0&	\cellcolor{First}62.7&	\cellcolor{Second}43.2&	\cellcolor{Third}70.7&	\cellcolor{Second}81.6\\
    \bottomrule
\end{tabular}%
\caption{Quantitative comparisons with SOTA methods on human local body poses. (PA-M.: PA-MPJPE, M.: MPJPE.)}
  \label{tab:localpose}%
\end{table}%
\begin{figure}[htbp]
  \centering
   \includegraphics[width=0.76\linewidth]{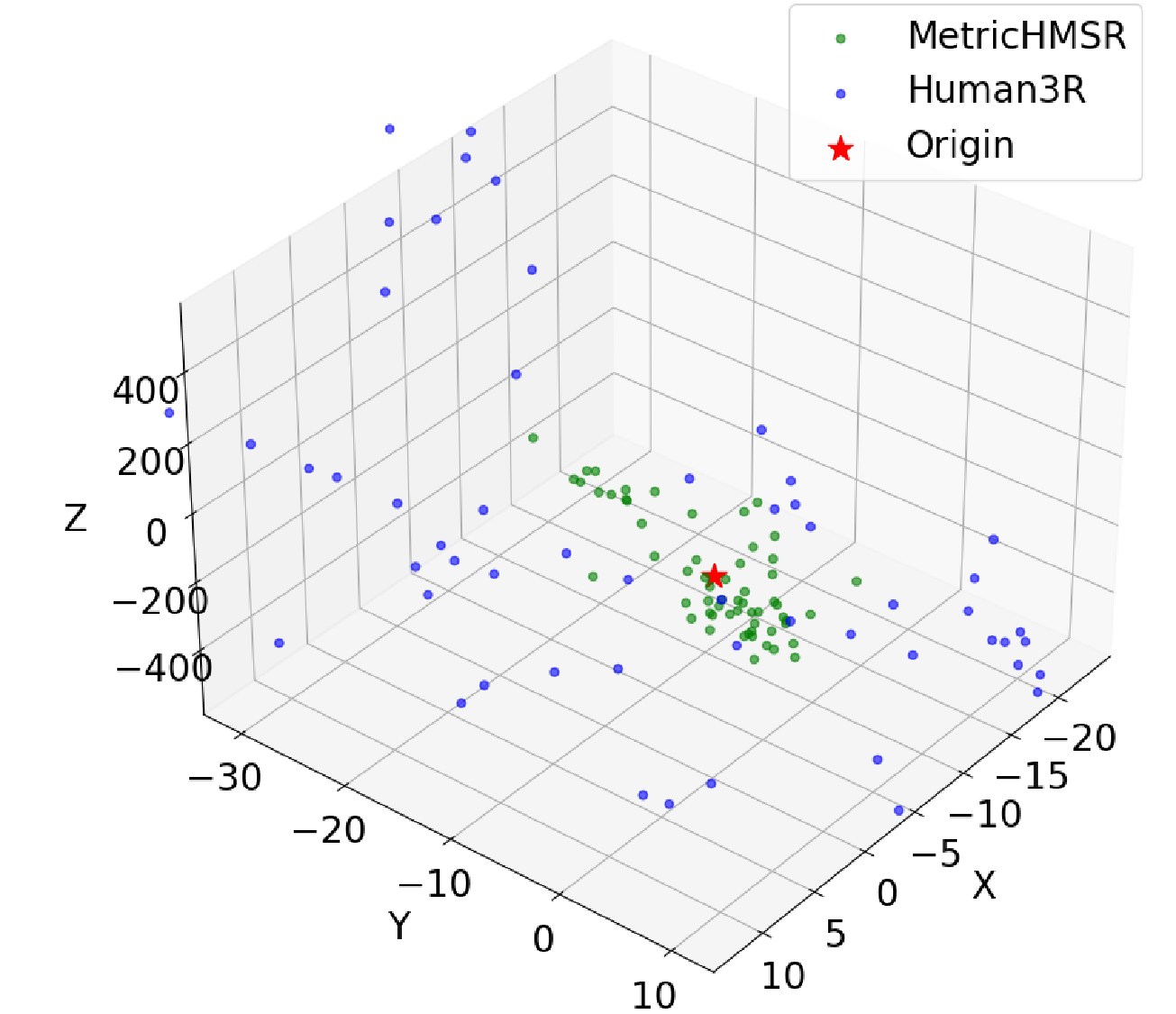}
   \caption{Distribution of normalized SMPL root joints in camera coordinates. Green points indicate results from MetricHMSR, while blue points correspond to Human3R results.
   }
   \label{fig:focal_root_distribution}
\end{figure}
\paragraph{Human mesh recovery under varying intrinsics.} 
Existing benchmarks typically assume fixed camera intrinsics, making it difficult to evaluate robustness to focal length variations. To address this limitation, we construct SynFocal, a synthetic dataset of human images rendered with varying focal lengths (see Supp. Mat.).
\cref{fig:focal_root_distribution} visualizes the distribution of normalized SMPL root joints. Our predictions remain consistently closer to the ground truth.

\paragraph{Metric Depth Estimation Comparison} We evaluate the performance of our depth refinement module using standard depth-estimation metrics: AbsRel (absolute relative error: $|d^{*}-d|/d$), MAE (mean absolute error: $|d^{*}-d|$), and $\delta_1$ (the percentage of pixels satisfying $\max(d/d^{*},\, d^{*}/d) < 1.25$).

\cref{tab:depth_prox} presents the results of MMDE. 
By incorporating human body information, our method enhances performance demonstrates superior performance.

\begin{figure}
  \centering
  \includegraphics[width=0.95\linewidth]{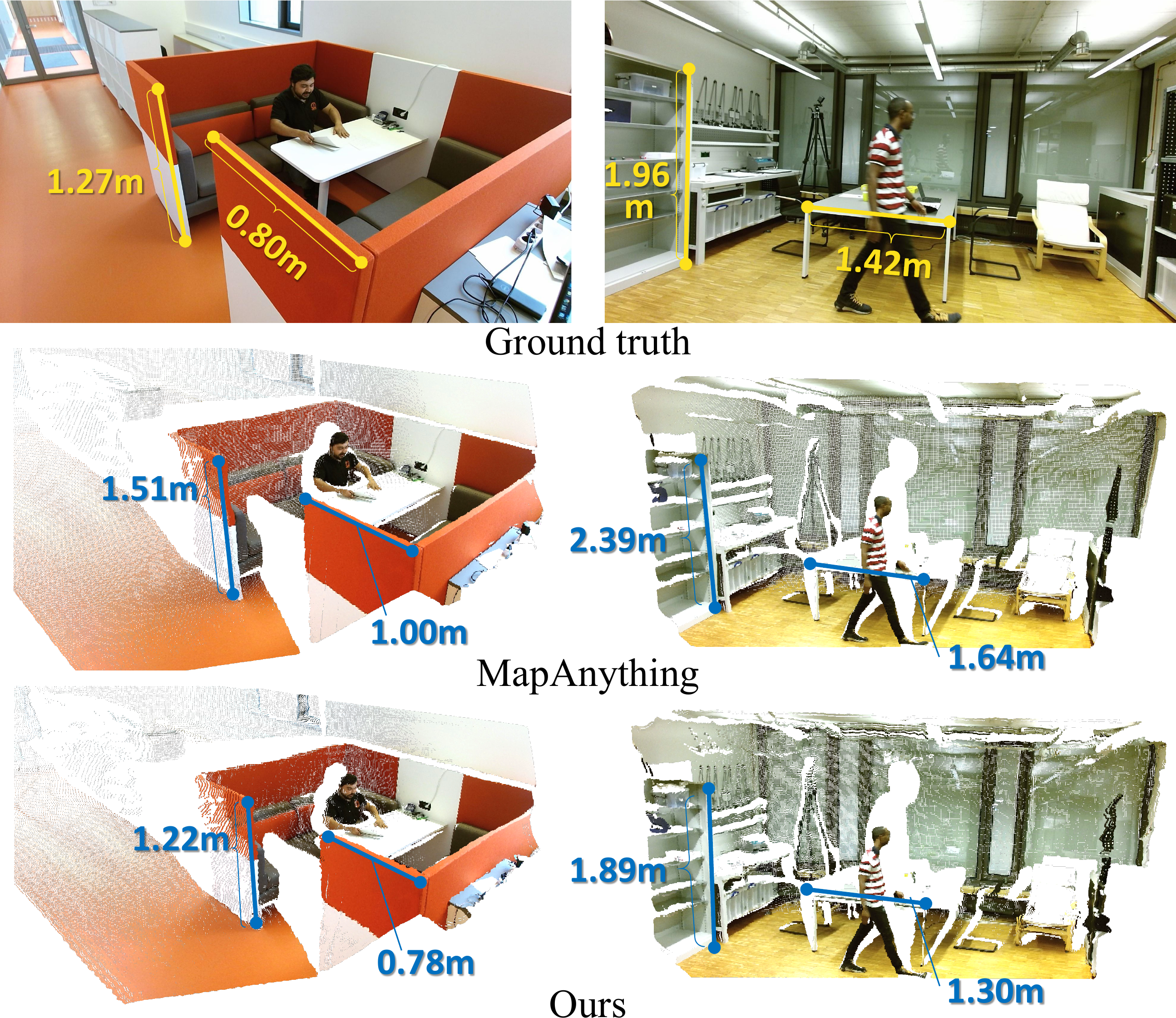}
  \caption{Comparison of the metric measurement on PROX.}
  \label{fig:measurement}
\end{figure}

\begin{table}[htbp]
  \centering
  \begin{tabular}{lccc}
    \toprule
    Method & AbsRel $\downarrow$ & MAE $\downarrow$ & $\delta_1$ $\uparrow$ \\
    \midrule
    Metric3D~\cite{yin2023metric3d}    & 0.76 & 2.40 & 0.01 \\
    Unidepth~\cite{piccinelli2024unidepth}    & 0.24 & 0.73 & 0.56  \\
    MapAnything~\cite{keetha2025mapanything} & 0.18 & 0.58 & 0.83  \\
    Ours        & \textbf{0.13} & \textbf{0.46} & \textbf{0.91} \\
    \bottomrule
  \end{tabular}
  \caption{Depth estimation comparison on PROX dataset.}
  \label{tab:depth_prox}
\end{table}
\cref{fig:measurement} presents a comparison of metric measurement results, demonstrating that our method achieves higher measurement accuracy.
\subsection{Ablation Study}
We validate the effectiveness of the key building blocks. As shown in \cref{tab:ablation}, the ablation study shows that our framework effectively exploits pixel-level camera ray information for 3D position estimation. Under a similar model size, replacing a standard transformer with the HumanMoE yields further gains. Moreover, both the Global MoE only and Patch MoE only variants underperform the full HumanMoE, suggesting that the two designs are complementary. We further study the impact of the number of route experts (the full + Ray \& MoE model uses 4 route experts).


\begin{table}[htbp]
\setlength{\tabcolsep}{1mm}
\centering
\begin{tabular}{lccccc}
    \toprule
    & \multicolumn{3}{c}{3DPW} & \multicolumn{2}{c}{EMDB-2}\\
    \cmidrule{2-6}
     & PA-M. & MPJPE & PVE & WA-M. & W-M. \\
    \midrule
    Image only & 35.6 & 57.2 & 66.8 & 64.5 & 191.8 \\
    + Ray & 34.4 & 55.0 & 64.9 & 56.4 & 154.1 \\
    + Ray \& MoE & \textbf{33.6} & \textbf{53.0} & \textbf{62.7} & 55.6 & \textbf{152.5} \\
    \midrule
    Global MoE only & 34.1 & 54.1 & 63.9 & 56.0 & 157.9 \\
    Patch MoE only & 34.2 & 54.3 & 64.0 & 55.6 & 154.7 \\
    \midrule
    0 Route Experts & 34.6 & 54.4 & 64.3 & 60.8 & 162.9 \\
    2 Route Experts & 34.3 & 53.8 & 63.5 & 55.7 & 153.0 \\
    8 Route Experts & 33.9 & 53.7 & 63.4 & \textbf{55.4} & 152.8 \\
    32 Route Experts & 34.2 & 53.9 & 63.8 & 56.3 & 154.8 \\
    \bottomrule
\end{tabular}
\caption{Ablation study of the key components of our method and additional ablation study on HumanMoE. (PA-M.: PA-MPJPE, WA-M.: WA-MPJPE, W-M.: W-MPJPE.)}
\label{tab:ablation}
\end{table}
\section{Conclusion}\label{sec:con}
In this paper, we propose a novel framework, MetricHMSR,  to regress human pose, metric shape, global 3D position as well as metric 3D scene from a single image. MetricHMSR unified the representation of camera projection model and human bounding box as the pixel-aligned camera ray map for encoding perspective 3D information. An specific designed HumanMoE network based on Mixture-of-Experts structure also proposed for more accurate and robust metric human mesh recovery.
We refined the monocular depth estimation given the metric human mesh, which further recovering the corresponding 3D scene in metric scale. 
Comprehensive experiments demonstrate that MetricHMSR achieves the best overall performance compared with existing online (single-frame input) methods. With MetricHMSR, it becomes possible to obtain metric pseudo-ground truth for in-the-wild public datasets. 

\noindent\textbf{Limitations and future work:} Our method currently does not support multi-person interaction perception and may struggle with severe occlusion. 
We have extended in-the-wild image datasets such as COCO, MPII, and AIC with generated pseudo-GT human meshes and 3D scenes. 
In future work, we plan to extend this annotation pipeline to large-scale internet data and investigate how such annotations can improve the performance and generalization of 3D human–scene reconstruction.

\noindent\textbf{Acknowledgments:} This work was supported in part by the NSFC (No.62422110 and No.82441013), the National Key R\&D Program of China (No.2024YFB2809101), the fellowship of China National Postdoctoral Program for Innovative Talents (BX20250399), Shuimu Tsinghua Scholar Program, and the Tsinghua University-Fuzhou Joint Institute for Data Technology (No. JIDT2024003).


{
    \small
    \bibliographystyle{ieeenat_fullname}
    \bibliography{main}

@inproceedings{dwivedi2024tokenhmr,
  title={Tokenhmr: Advancing human mesh recovery with a tokenized pose representation},
  author={Dwivedi, Sai Kumar and Sun, Yu and Patel, Priyanka and Feng, Yao and Black, Michael J},
  booktitle={Proceedings of the IEEE/CVF Conference on Computer Vision and Pattern Recognition},
  pages={1323--1333},
  year={2024}
}

@article{sarandi2025neural,
  title={Neural localizer fields for continuous 3d human pose and shape estimation},
  author={S{\'a}r{\'a}ndi, Istv{\'a}n and Pons-Moll, Gerard},
  journal={Advances in Neural Information Processing Systems},
  volume={37},
  pages={140032--140065},
  year={2025}
}

@inproceedings{li2022cliff,
  title={Cliff: Carrying location information in full frames into human pose and shape estimation},
  author={Li, Zhihao and Liu, Jianzhuang and Zhang, Zhensong and Xu, Songcen and Yan, Youliang},
  booktitle={Proceedings of the European Conference on Computer Vision},
  pages={590--606},
  year={2022}
}

@inproceedings{kocabas2020vibe,
  title={VIBE: Video Inference for Human Body Pose and Shape Estimation},
  author={Kocabas, Muhammed and Athanasiou, Nikos and Black, Michael J.},
  booktitle={Proceedings of the IEEE/CVF Conference on Computer Vision and Pattern Recognition},
  month = {June},
  pages={5252-5262},
  year = {2020},
}

@inproceedings{kolotouros2019spin,
  title={Learning to reconstruct 3D human pose and shape via model-fitting in the loop},
  author={Kolotouros, Nikos and Pavlakos, Georgios and Black, Michael J and Daniilidis, Kostas},
  booktitle={Proceedings of the IEEE International Conference on Computer Vision},
  pages={2252--2261},
  year={2019}
}

@inproceedings{goel2023hmrv2,
  title={Humans in 4D: Reconstructing and tracking humans with transformers},
  author={Goel, Shubham and Pavlakos, Georgios and Rajasegaran, Jathushan and Kanazawa, Angjoo and Malik, Jitendra},
  booktitle={Proceedings of the IEEE/CVF International Conference on Computer Vision},
  pages={14783--14794},
  year={2023}
}

@inproceedings{bogo2016smplify,
  title={Keep it SMPL: Automatic Estimation of 3D Human Pose and Shape from a Single Image},
  author={Bogo, Federica and Kanazawa, Angjoo and Lassner, Christoph and Gehler, Peter and Romero, Javier and Black, Michael J.},
  booktitle={Proceedings of the European Conference on Computer Vision},
  pages={561--578},
  year={2016}
}

@inproceedings{pavlakos2019smplx,
  title={Expressive body capture: 3d hands, face, and body from a single image},
  author={Pavlakos, Georgios and Choutas, Vasileios and Ghorbani, Nima and Bolkart, Timo and Osman, Ahmed AA and Tzionas, Dimitrios and Black, Michael J},
  booktitle={Proceedings of the IEEE/CVF Conference on Computer Vision and Pattern Recognition},
  pages={10975--10985},
  year={2019}
}

@inproceedings{zhang2021pymaf,
  title={Pymaf: 3d human pose and shape regression with pyramidal mesh alignment feedback loop},
  author={Zhang, Hongwen and Tian, Yating and Zhou, Xinchi and Ouyang, Wanli and Liu, Yebin and Wang, Limin and Sun, Zhenan},
  booktitle={Proceedings of the IEEE/CVF International Conference on Computer Vision},
  pages={11446--11456},
  year={2021}
}

@article{zhang2023pymafx,
  title={Pymaf-x: Towards well-aligned full-body model regression from monocular images},
  author={Zhang, Hongwen and Tian, Yating and Zhang, Yuxiang and Li, Mengcheng and An, Liang and Sun, Zhenan and Liu, Yebin},
  journal={IEEE Transactions on Pattern Analysis and Machine Intelligence},
  volume={45},
  number={10},
  pages={12287--12303},
  year={2023},
  publisher={IEEE}
}

@article{cai2024smpler,
  title={Smpler-x: Scaling up expressive human pose and shape estimation},
  author={Cai, Zhongang and Yin, Wanqi and Zeng, Ailing and Wei, Chen and Sun, Qingping and Yanjun, Wang and Pang, Hui En and Mei, Haiyi and Zhang, Mingyuan and Zhang, Lei and others},
  journal={Advances in Neural Information Processing Systems},
  volume={36},
  year={2024}
}

@inproceedings{kanazawa2018hmr,
  title={End-to-end recovery of human shape and pose},
  author={Kanazawa, Angjoo and Black, Michael J and Jacobs, David W and Malik, Jitendra},
  booktitle={Proceedings of the IEEE/CVF Conference on Computer Vision and Pattern Recognition},
  pages={7122--7131},
  year={2018}
}

@article{loper2015smpl,
    title={SMPL: A skinned multi-person linear model},
    author={Loper, Matthew and Mahmood, Naureen and Romero, Javier and Pons-Moll, Gerard and Black, Michael J},
    journal={ACM Transactions on Graphics},
    volume={34},
    number={6},
    pages={1--16},
    year={2015},
    publisher={ACM New York, NY, USA}
}

@inproceedings{wang2023zolly,
  title={Zolly: Zoom focal length correctly for perspective-distorted human mesh reconstruction},
  author={Wang, Wenjia and Ge, Yongtao and Mei, Haiyi and Cai, Zhongang and Sun, Qingping and Wang, Yanjun and Shen, Chunhua and Yang, Lei and Komura, Taku},
  booktitle={Proceedings of the IEEE/CVF International Conference on Computer Vision},
  pages={3925--3935},
  year={2023}
}

@inproceedings{sun2022bve,
  title={Putting people in their place: Monocular regression of 3d people in depth},
  author={Sun, Yu and Liu, Wu and Bao, Qian and Fu, Yili and Mei, Tao and Black, Michael J},
  booktitle={Proceedings of the IEEE/CVF Conference on Computer Vision and Pattern Recognition},
  pages={13243--13252},
  year={2022}
}

@inproceedings{sun2023trace,
  title={Trace: 5d temporal regression of avatars with dynamic cameras in 3d environments},
  author={Sun, Yu and Bao, Qian and Liu, Wu and Mei, Tao and Black, Michael J},
  booktitle={Proceedings of the IEEE/CVF Conference on Computer Vision and Pattern Recognition},
  pages={8856--8866},
  year={2023}
}

@inproceedings{shin2024wham,
  title={Wham: Reconstructing world-grounded humans with accurate 3d motion},
  author={Shin, Soyong and Kim, Juyong and Halilaj, Eni and Black, Michael J},
  booktitle={Proceedings of the IEEE/CVF Conference on Computer Vision and Pattern Recognition},
  pages={2070--2080},
  year={2024}
}

@inproceedings{wang2024tram,
  title={TRAM: Global Trajectory and Motion of 3D Humans from in-the-wild Videos},
  author={Wang, Yufu and Wang, Ziyun and Liu, Lingjie and Daniilidis, Kostas},
  booktitle={Proceedings of the European Conference on Computer Vision},
  pages={467--487},
  year={2024},
  organization={Springer}
}

@article{wang2024blade,
  title={BLADE: Single-view Body Mesh Learning through Accurate Depth Estimation},
  author={Wang, Shengze and Li, Jiefeng and Li, Tianye and Yuan, Ye and Fuchs, Henry and Nagano, Koki and De Mello, Shalini and Stengel, Michael},
  journal={arXiv preprint arXiv:2412.08640},
  year={2024}
}

@article{pham2024sharpdepth,
  title={SharpDepth: Sharpening Metric Depth Predictions Using Diffusion Distillation},
  author={Pham, Duc-Hai and Do, Tung and Nguyen, Phong and Hua, Binh-Son and Nguyen, Khoi and Nguyen, Rang},
  journal={arXiv preprint arXiv:2411.18229},
  year={2024}
}

@inproceedings{yin2023metric3d,
  title={Metric3d: Towards zero-shot metric 3d prediction from a single image},
  author={Yin, Wei and Zhang, Chi and Chen, Hao and Cai, Zhipeng and Yu, Gang and Wang, Kaixuan and Chen, Xiaozhi and Shen, Chunhua},
  booktitle={Proceedings of the IEEE/CVF International Conference on Computer Vision},
  pages={9043--9053},
  year={2023}
}

@inproceedings{piccinelli2024unidepth,
  title={UniDepth: Universal monocular metric depth estimation},
  author={Piccinelli, Luigi and Yang, Yung-Hsu and Sakaridis, Christos and Segu, Mattia and Li, Siyuan and Van Gool, Luc and Yu, Fisher},
  booktitle={Proceedings of the IEEE/CVF Conference on Computer Vision and Pattern Recognition},
  pages={10106--10116},
  year={2024}
}

@article{yang2024dav2,
  title={Depth anything v2},
  author={Yang, Lihe and Kang, Bingyi and Huang, Zilong and Zhao, Zhen and Xu, Xiaogang and Feng, Jiashi and Zhao, Hengshuang},
  journal={Advances in Neural Information Processing Systems},
  volume={37},
  pages={21875--21911},
  year={2024}
}

@inproceedings{andriluka2014mpii,
  title={2d human pose estimation: New benchmark and state of the art analysis},
  author={Andriluka, Mykhaylo and Pishchulin, Leonid and Gehler, Peter and Schiele, Bernt},
  booktitle={Proceedings of the IEEE/CVF Conference on Computer Vision and Pattern Recognition},
  pages={3686--3693},
  year={2014}
}

@inproceedings{lin2014coco,
  title={Microsoft coco: Common objects in context},
  author={Lin, Tsung-Yi and Maire, Michael and Belongie, Serge and Hays, James and Perona, Pietro and Ramanan, Deva and Doll{\'a}r, Piotr and Zitnick, C Lawrence},
  booktitle={Proceedings of the European Conference on Computer Vision},
  pages={740--755},
  year={2014},
  organization={Springer}
}

@article{bhat2023zoedepth,
  title={Zoedepth: Zero-shot transfer by combining relative and metric depth},
  author={Bhat, Shariq Farooq and Birkl, Reiner and Wofk, Diana and Wonka, Peter and M{\"u}ller, Matthias},
  journal={arXiv preprint arXiv:2302.12288},
  year={2023}
}

@inproceedings{zhao2024metric,
  title={Metric from human: Zero-shot monocular metric depth estimation via test-time adaptation},
  author={Zhao, Yizhou and Bian, Hengwei and Chen, Kaihua and Ji, Pengliang and Qu, Liao and Lin, Shao-yu and Yu, Weichen and Li, Haoran and Chen, Hao and Shen, Jun and others},
  booktitle={The Thirty-eighth Annual Conference on Neural Information Processing Systems},
  year={2024}
}

@inproceedings{Mueller2021tuch,
  title = {On Self-Contact and Human Pose},
  author = {M{\"u}ller, Lea and Osman, Ahmed A. A. and Tang, Siyu and Huang, Chun-Hao P. and Black, Michael J.},
  booktitle = {Proceedings of the IEEE/CVF Conference on Computer Vision and Pattern Recognition},
  month = jun,
  year = {2021},
  pages={9990-9999},
  month_numeric = {6}
}

@inproceedings{shen2024gvhmr,
  title={World-Grounded Human Motion Recovery via Gravity-View Coordinates},
  author={Shen, Zehong and Pi, Huaijin and Xia, Yan and Cen, Zhi and Peng, Sida and Hu, Zechen and Bao, Hujun and Hu, Ruizhen and Zhou, Xiaowei},
  booktitle={SIGGRAPH Asia 2024 Conference Papers},
  pages={1--11},
  year={2024}
}

@inproceedings{von20183dpw,
  title={Recovering accurate 3d human pose in the wild using imus and a moving camera},
  author={Von Marcard, Timo and Henschel, Roberto and Black, Michael J and Rosenhahn, Bodo and Pons-Moll, Gerard},
  booktitle={Proceedings of the European conference on computer vision},
  pages={601--617},
  year={2018}
}

@inproceedings{hassan2019prox,
  title={Resolving 3D human pose ambiguities with 3D scene constraints},
  author={Hassan, Mohamed and Choutas, Vasileios and Tzionas, Dimitrios and Black, Michael J},
  booktitle={Proceedings of the IEEE/CVF International Conference on Computer Vision},
  pages={2282--2292},
  year={2019}
}

@inproceedings{kaufmann2023emdb,
  title={Emdb: The electromagnetic database of global 3d human pose and shape in the wild},
  author={Kaufmann, Manuel and Song, Jie and Guo, Chen and Shen, Kaiyue and Jiang, Tianjian and Tang, Chengcheng and Z{\'a}rate, Juan Jos{\'e} and Hilliges, Otmar},
  booktitle={Proceedings of the IEEE/CVF International Conference on Computer Vision},
  pages={14632--14643},
  year={2023}
}

@article{teed2021droid,
  title={Droid-slam: Deep visual slam for monocular, stereo, and rgb-d cameras},
  author={Teed, Zachary and Deng, Jia},
  journal={Advances in neural information processing systems},
  volume={34},
  pages={16558--16569},
  year={2021}
}

@inproceedings{ye2023slamhmr,
  title={Decoupling human and camera motion from videos in the wild},
  author={Ye, Vickie and Pavlakos, Georgios and Malik, Jitendra and Kanazawa, Angjoo},
  booktitle={Proceedings of the IEEE/CVF Conference on Computer Vision and Pattern Recognition},
  pages={21222--21232},
  year={2023}
}

@inproceedings{kocabas2021spec,
  title={SPEC: Seeing people in the wild with an estimated camera},
  author={Kocabas, Muhammed and Huang, Chun-Hao P and Tesch, Joachim and M{\"u}ller, Lea and Hilliges, Otmar and Black, Michael J},
  booktitle={Proceedings of the IEEE/CVF International Conference on Computer Vision},
  pages={11035--11045},
  year={2021}
}

@inproceedings{Baradel2024multihmr,
    title={Multi-HMR: Multi-Person Whole-Body Human Mesh Recovery in a Single Shot},
    author={Baradel, Fabien and 
            Armando, Matthieu and 
            Galaaoui, Salma and 
            Br{\'e}gier, Romain and 
            Weinzaepfel, Philippe and 
            Rogez, Gr{\'e}gory and
            Lucas, Thomas
            },
    booktitle={Proceedings of the European conference on computer vision},
    year={2024}
}

@inproceedings{patel2024camerahmr, 
    title={CameraHMR: Aligning People with Perspective}, 
    author={Patel, Priyanka and Black, Michael J.}, 
    booktitle={2025 International Conference on 3D Vision}, 
    year={2025} 
}

@inproceedings{dosovitskiy2021vit,
  title     = {An Image is Worth 16×16 Words: Transformers for Image Recognition at Scale},
  author    = {Alexey Dosovitskiy and Lucas Beyer and Alexander Kolesnikov and Dirk Weissenborn and Xiaohua Zhai and Thomas Unterthiner and Mostafa Dehghani and Matthias Minderer and Georg Heigold and Sylvain Gelly and Jakob Uszkoreit and Neil Houlsby},
  booktitle = {International Conference on Learning Representations},
  year      = {2021}
}

@INPROCEEDINGS{Wang2025vggt,
  author={Wang, Jianyuan and Chen, Minghao and Karaev, Nikita and Vedaldi, Andrea and Rupprecht, Christian and Novotny, David},
  booktitle={Proceedings of the IEEE/CVF Conference on Computer Vision and Pattern Recognition}, 
  title={VGGT: Visual Geometry Grounded Transformer}, 
  year={2025},
  pages={5294-5306}}

@article{mildenhall2021nerf,
  title={Nerf: Representing scenes as neural radiance fields for view synthesis},
  author={Mildenhall, Ben and Srinivasan, Pratul P and Tancik, Matthew and Barron, Jonathan T and Ramamoorthi, Ravi and Ng, Ren},
  journal={Communications of the ACM},
  volume={65},
  number={1},
  pages={99--106},
  year={2021},
  publisher={ACM New York, NY, USA}
}

@inproceedings{dai2024deepseekmoe,
  title={Deepseekmoe: Towards ultimate expert specialization in mixture-of-experts language models},
  author={Dai, Damai and Deng, Chengqi and Zhao, Chenggang and Xu, RX and Gao, Huazuo and Chen, Deli and Li, Jiashi and Zeng, Wangding and Yu, Xingkai and Wu, Yu and others},
  booktitle={Proceedings of the 62nd Annual Meeting of the Association for Computational Linguistics (Volume 1: Long Papers)},
  pages={1280--1297},
  year={2024}
}

@inproceedings{kissos2020beyond,
  title={Beyond weak perspective for monocular 3d human pose estimation},
  author={Kissos, Imry and Fritz, Lior and Goldman, Matan and Meir, Omer and Oks, Eduard and Kliger, Mark},
  booktitle={Proceedings of the European Conference on Computer Vision},
  pages={541--554},
  year={2020},
  organization={Springer}
}

@inproceedings{jang2025pow3r,
  title={Pow3r: Empowering unconstrained 3d reconstruction with camera and scene priors},
  author={Jang, Wonbong and Weinzaepfel, Philippe and Leroy, Vincent and Agapito, Lourdes and Revaud, Jerome},
  booktitle={Proceedings of the IEEE/CVF Conference on Computer Vision and Pattern Recognition},
  pages={1071--1081},
  year={2025}
}

@article{xu2022vitpose,
  title={Vitpose: Simple vision transformer baselines for human pose estimation},
  author={Xu, Yufei and Zhang, Jing and Zhang, Qiming and Tao, Dacheng},
  journal={Advances in neural information processing systems},
  volume={35},
  pages={38571--38584},
  year={2022}
}

@article{muqeeth2023soft,
  title={Soft merging of experts with adaptive routing},
  author={Muqeeth, Mohammed and Liu, Haokun and Raffel, Colin},
  journal={arXiv preprint arXiv:2306.03745},
  year={2023}
}

@inproceedings{zeng2022not,
  title={Not all tokens are equal: Human-centric visual analysis via token clustering transformer},
  author={Zeng, Wang and Jin, Sheng and Liu, Wentao and Qian, Chen and Luo, Ping and Ouyang, Wanli and Wang, Xiaogang},
  booktitle={Proceedings of the IEEE/CVF Conference on Computer Vision and Pattern Recognition},
  pages={11101--11111},
  year={2022}
}

@inproceedings{wang2025cut3r,
  title={Continuous 3D Perception Model with Persistent State},
  author={Wang, Qianqian and Zhang, Yifei and Holynski, Aleksander and Efros, Alexei A and Kanazawa, Angjoo},
  booktitle={Proceedings of the IEEE/CVF Conference on Computer Vision and Pattern Recognition},
  pages={10510--10522},
  year={2025}
}

@article{chen2025human3r,
    title={Human3R: Everyone Everywhere All at Once},
    author={Chen, Yue and Chen, Xingyu and Xue, Yuxuan and Chen, Anpei and Xiu, Yuliang and Gerard, Pons-Moll},
    journal={arXiv preprint arXiv:2510.06219},
    year={2025}
    }

@article{liu2024deepseekv2,
  title={Deepseek-v2: A strong, economical, and efficient mixture-of-experts language model},
  author={Liu, Aixin and Feng, Bei and Wang, Bin and Wang, Bingxuan and Liu, Bo and Zhao, Chenggang and Dengr, Chengqi and Ruan, Chong and Dai, Damai and Guo, Daya and others},
  journal={arXiv preprint arXiv:2405.04434},
  year={2024}
}

@article{han2024vimoe,
  title={Vimoe: An empirical study of designing vision mixture-of-experts},
  author={Han, Xumeng and Wei, Longhui and Dou, Zhiyang and Wang, Zipeng and Qiang, Chenhui and He, Xin and Sun, Yingfei and Han, Zhenjun and Tian, Qi},
  journal={arXiv preprint arXiv:2410.15732},
  year={2024}
}

@inproceedings{fu20213dfront,
  title={3D-FRONT: 3d furnished rooms with layouts and semantics},
  author={Fu, Huan and Cai, Bowen and Gao, Lin and Zhang, Ling-Xiao and Wang, Jiaming and Li, Cao and Zeng, Qixun and Sun, Chengyue and Jia, Rongfei and Zhao, Binqiang and others},
  booktitle={Proceedings of the IEEE/CVF International Conference on Computer Vision},
  pages={10933--10942},
  year={2021}
}

@inproceedings{yu2021function4d,
  title={Function4d: Real-time human volumetric capture from very sparse consumer rgbd sensors},
  author={Yu, Tao and Zheng, Zerong and Guo, Kaiwen and Liu, Pengpeng and Dai, Qionghai and Liu, Yebin},
  booktitle={Proceedings of the IEEE/CVF Conference on Computer Vision and Pattern Recognition},
  pages={5746--5756},
  year={2021}
}

@misc{blender,
    key = {Blender},
    year={2025},
    note = {https://www.blender.org}
}

@inproceedings{hao2025perspose,
  title={PersPose: 3D Human Pose Estimation with Perspective Encoding and Perspective Rotation},
  author={Hao, Xiaoyang and Li, Han},
  booktitle={Proceedings of the IEEE/CVF International Conference on Computer Vision},
  pages={8110--8119},
  year={2025}
}

@article{keetha2025mapanything,
  title={MapAnything: Universal feed-forward metric 3D reconstruction},
  author={Keetha, Nikhil and M{\"u}ller, Norman and Sch{\"o}nberger, Johannes and Porzi, Lorenzo and Zhang, Yuchen and Fischer, Tobias and Knapitsch, Arno and Zauss, Duncan and Weber, Ethan and Antunes, Nelson and others},
  journal={arXiv preprint arXiv:2509.13414},
  year={2025}
}

@inproceedings{li2024coin,
  title={Coin: Control-inpainting diffusion prior for human and camera motion estimation},
  author={Li, Jiefeng and Yuan, Ye and Rempe, Davis and Zhang, Haotian and Molchanov, Pavlo and Lu, Cewu and Kautz, Jan and Iqbal, Umar},
  booktitle={Proceedings of the European Conference on Computer Vision},
  pages={426--446},
  year={2024},
  organization={Springer}
}

@inproceedings{yuan2022glamr,
  title={Glamr: Global occlusion-aware human mesh recovery with dynamic cameras},
  author={Yuan, Ye and Iqbal, Umar and Molchanov, Pavlo and Kitani, Kris and Kautz, Jan},
  booktitle={Proceedings of the IEEE/CVF Conference on Computer Vision and Pattern Recognition},
  pages={11038--11049},
  year={2022}
}

@inproceedings{wang2025prompthmr,
  title={PromptHMR: Promptable Human Mesh Recovery},
  author={Wang, Yufu and Sun, Yu and Patel, Priyanka and Daniilidis, Kostas and Black, Michael J and Kocabas, Muhammed},
  booktitle={Proceedings of the IEEE/CVF Conference on Computer Vision and Pattern Recognition},
  pages={1148--1159},
  year={2025}
}

@inproceedings{black2023bedlam,
  title={Bedlam: A synthetic dataset of bodies exhibiting detailed lifelike animated motion},
  author={Black, Michael J and Patel, Priyanka and Tesch, Joachim and Yang, Jinlong},
  booktitle={Proceedings of the IEEE/CVF Conference on Computer Vision and Pattern Recognition},
  pages={8726--8737},
  year={2023}
}

@article{wu2017ai,
  title={Ai challenger: A large-scale dataset for going deeper in image understanding},
  author={Wu, Jiahong and Zheng, He and Zhao, Bo and Li, Yixin and Yan, Baoming and Liang, Rui and Wang, Wenjia and Zhou, Shipei and Lin, Guosen and Fu, Yanwei and others},
  journal={arXiv preprint arXiv:1711.06475},
  year={2017}
}

@inproceedings{tesch2025bedlam2,
  title={{BEDLAM}2.0: Synthetic humans and cameras in motion},
  author={Joachim Tesch and Giorgio Becherini and Prerana Achar and Anastasios Yiannakidis and Muhammed Kocabas and Priyanka Patel and Michael J. Black},
  booktitle={The Thirty-ninth Annual Conference on Neural Information Processing Systems Datasets and Benchmarks Track},
  year={2025}
}

@article{shazeer2017outrageously,
  title={Outrageously large neural networks: The sparsely-gated mixture-of-experts layer},
  author={Shazeer, Noam and Mirhoseini, Azalia and Maziarz, Krzysztof and Davis, Andy and Le, Quoc and Hinton, Geoffrey and Dean, Jeff},
  journal={arXiv preprint arXiv:1701.06538},
  year={2017}
}

@article{fedus2022switch,
  title={Switch transformers: Scaling to trillion parameter models with simple and efficient sparsity},
  author={Fedus, William and Zoph, Barret and Shazeer, Noam},
  journal={Journal of Machine Learning Research},
  volume={23},
  number={120},
  pages={1--39},
  year={2022}
}

@inproceedings{choutas2022accurate,
  title={Accurate 3D body shape regression using metric and semantic attributes},
  author={Choutas, Vasileios and M{\"u}ller, Lea and Huang, Chun-Hao P and Tang, Siyu and Tzionas, Dimitrios and Black, Michael J},
  booktitle={Proceedings of the IEEE/CVF Conference on Computer Vision and Pattern Recognition},
  pages={2718--2728},
  year={2022}
}

@inproceedings{tirado2025anycalib,
  title={AnyCalib: On-manifold learning for model-agnostic single-view camera calibration},
  author={Tirado-Gar{\'\i}n, Javier and Civera, Javier},
  booktitle={Proceedings of the IEEE/CVF International Conference on Computer Vision},
  pages={8044--8055},
  year={2025}
}
}

\clearpage
\setcounter{page}{1}
\maketitlesupplementary

\section{More Implementation Details}
\subsection{Camera, Image and Metrics}\label{sec:bias}
In human mesh recovery, full perspective projection~\cite{li2022cliff, Baradel2024multihmr, wang2024tram,hao2025perspose} has gained increasing attention. The development of monocular metric depth estimation~\cite{bhat2023zoedepth, yin2023metric3d, zhao2024metric, goel2023hmrv2, piccinelli2024unidepth, pham2024sharpdepth, yang2024dav2} and single-image intrinsic parameter estimation~\cite{Wang2025vggt,keetha2025mapanything} has made it possible to recover metric human mesh from a single image. The intrinsic parameters of the camera and the bounding boxes (bbox) are crucial to perceiving human metric information and the 3D position.

\begin{figure}[htbp]
  \centering
   \includegraphics[width=\linewidth]{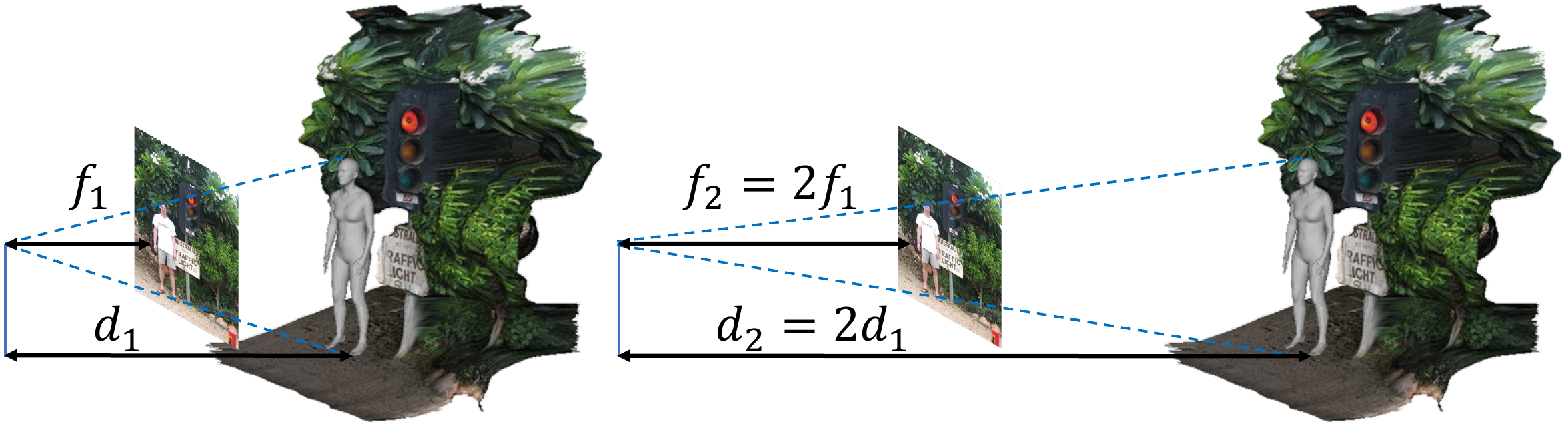}
   \caption{Illustration of the human body's metric distance under the same 2D projection with different focal lengths. Focal: $f$, distance: $d$.}
   \label{fig:3_focal}
\end{figure}

 \cref{fig:3_focal} shows the impact of the camera's intrinsic parameters on the metric depth of the human. Under the same resolution, if two cameras ($f_2 = 2f_1$) produce the same image size for the same person, their metric depths are different ($d_2 = 2d_1$). If the focal length can be provided as a known parameter to the network, it becomes possible for the network to infer the metric depth of human body.

\begin{figure}[htbp]
  \centering
   \includegraphics[width=0.65\linewidth]{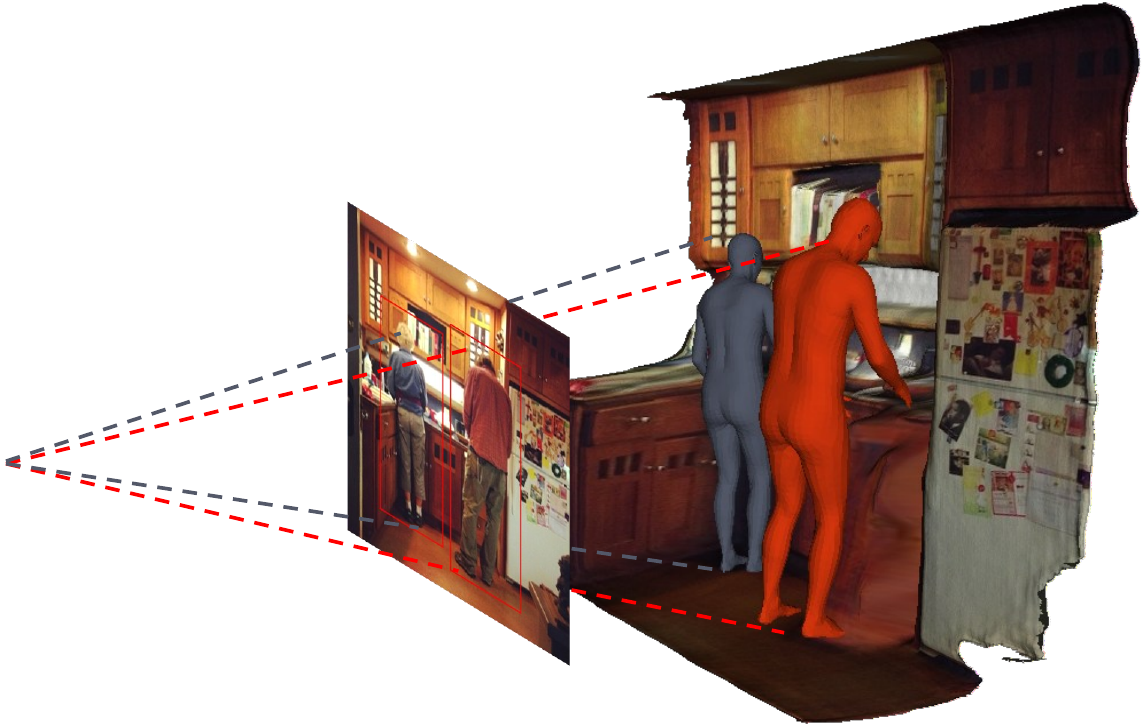}
   \caption{Illustration of the human body's metric position with different bounding box. The position of the bounding box directly influences the 3D location of the human body.}
   \label{fig:3_bbox}
\end{figure}

As shown in \cref{fig:3_bbox}, the distance between the bounding box and the principal point ($c_x$, $c_y$) is directly related to the 3D position of the human body in the camera space.

\subsection{Camera Ray Representation}
CLIFF~\cite{li2022cliff} was the first to recognize the global information contained in the bounding box, and it incorporated bounding-box as additional input to the network. However, this approach neglects the role of camera intrinsics. To address this limitation, we draw inspiration from NeRF~\cite{mildenhall2021nerf} and introduce the bounding ray map representation method, as shown in \cref{fig:4_camera_ray}.

\subsection{Losses of Metric Human Mesh Recovery}
In \cref{eq:hmr_losses}, each loss term is computed as

\begin{align}
    \begin{aligned}
    \mathcal{L}_{J_{2D}} =& ||\hat{J}_{2D}-J_{2D}||_2,\\
    \mathcal{L}_{J_{3D}} =& ||\hat{J}_{3D}-J_{3D}||_2,\\
    \mathcal{L}_{V_{3D}} =& ||\hat{V}_{3D}-V_{3D}||_2,\\
    \mathcal{L}_{\theta} =& ||\hat{\theta}-\theta||_2,\\
    \mathcal{L}_{\beta} =& ||\hat{\beta}-\beta||_2,\\
    \mathcal{L}_{h} =& ||\hat{h}-h||_2,\\
    \end{aligned}
\end{align}
where
$J_{2D}$, $J_{3D}$, $V_{3D}$, $\theta$, $\beta$, and $h$ represent the ground truth of the 2D keypoints, the 3D keypoints, the vertices of the SMPL model, the pose parameters of the SMPL and the shape parameters of the SMPL, respectively. 
$\hat{J}_{2D}$, $\hat{J}_{3D}$, $\hat{V}_{3D}$, $\hat{\theta}$, $\hat{\beta}$, and $\hat{h}$ represent the corresponding network predictions of the 2D keypoints, the 3D keypoints, the vertices of the SMPL model, the pose parameters of the SMPL and the shape parameters of the SMPL, respectively. 

\subsection{HumanMoE}
On one hand, HumanMoE condenses all 3D attributes into different experts within a single module. On the other hand, it assigns features from different hierarchies and dimensions to specialized experts. 

At the patch level, the Patch MoE routes patches of different body parts to distinct experts. As shown in \cref{fig:5_expert_heatmap}, this leads to semantic decoupling and feature-level disentanglement, enabling specialized processing and improving performance.

\begin{figure}
  \centering
   \includegraphics[width=0.96\linewidth]{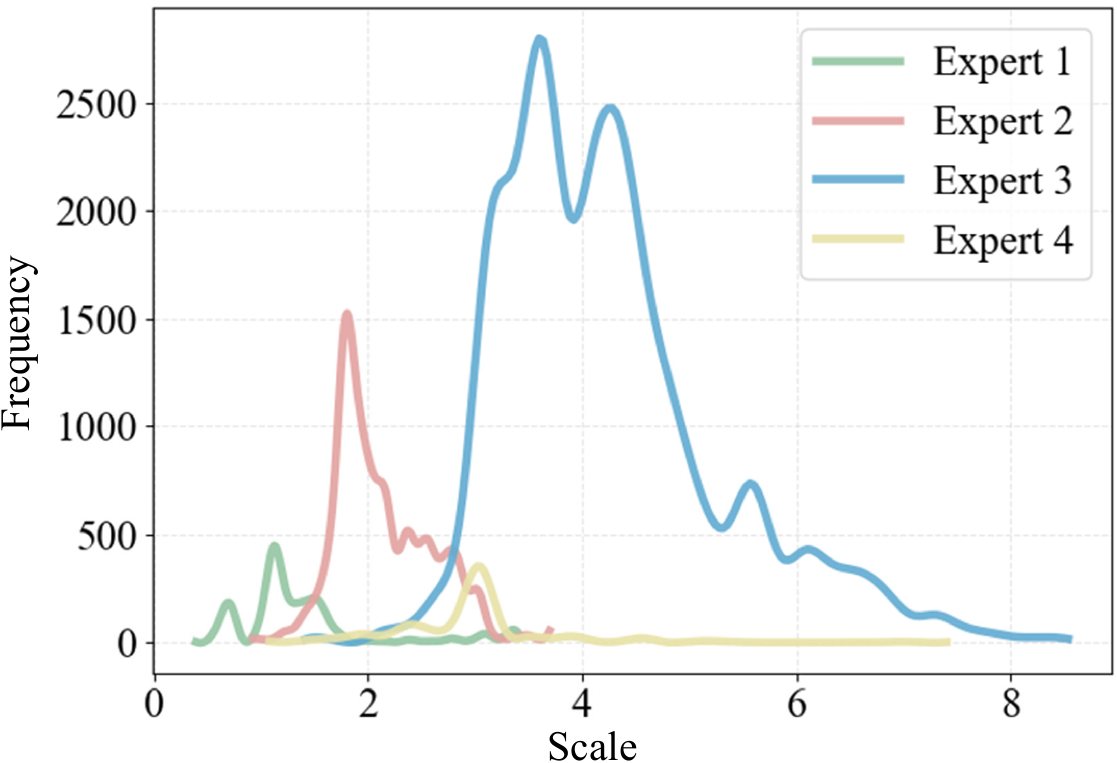}
   \caption{Expert allocation of the global MoE across different bbox sizes. Scale denotes the factor by which the cropped images are resized to a fixed size.}
   \label{fig:x_globalMoE_scale}
\end{figure}
\begin{figure}
  \centering
   \includegraphics[width=\linewidth]{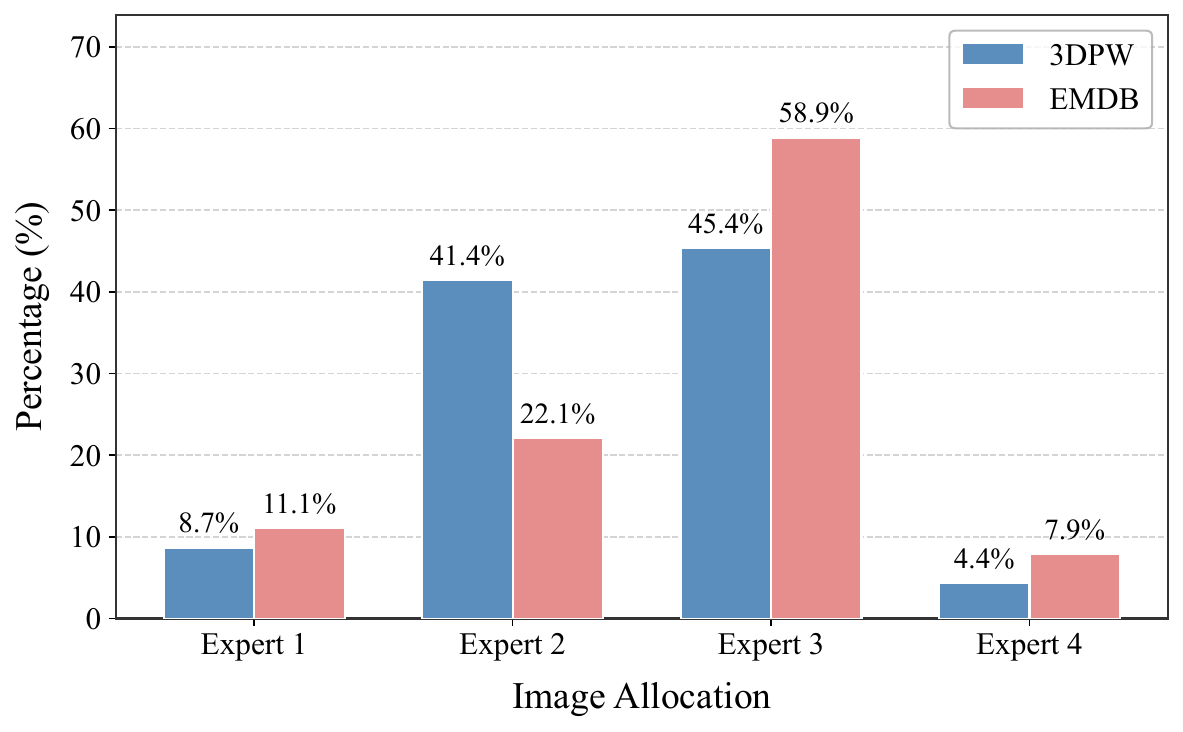}
   \caption{Expert allocation of the global MoE on different datasets.}
   \label{fig:x_globalMoE_dataset}
\end{figure}

At the global image level, the Global MoE routes the aggregated image feature to specialized experts via a single query token. Complementing the Patch MoE, it operates holistically to capture global context and enable consistent reasoning over metric properties for final prediction. \cref{fig:x_globalMoE_scale} and \cref{fig:x_globalMoE_dataset} illustrate its role.

We tested on the fixed-intrinsics dataset 3DPW. The cropped images are scaled to a fixed size of $256 \times 256$ to feed to our network. This scale reflects the size of the bounding box (bbox). \cref{fig:x_globalMoE_dataset} shows how the first layer of the global MoE allocates images in accordance with this scale. It can be observed that images with larger bounding boxes (i.e., closer people) tend to be assigned to Expert 3, while those with smaller bounding boxes (i.e., farther people) are more often assigned to Expert 1.

\cref{fig:x_globalMoE_dataset} shows the assignment of images from the 3DPW and EMDB datasets to experts in the final layer of the MoE. Due to differences in human pose, scene, and other factors across datasets, the allocation to experts also differs.

\subsection{{Human-Guided Metric Depth Refinement}}

We refine a predicted dense depth \(x_{\text{in}}\) into metric-consistent depth by learning per-pixel affine parameters \((s,b)\) so that \(x_{\text{out}}(u,v)=s(u,v)\,x_{\text{in}}(u,v)+b(u,v)\). Absolute-scale constraints (``anchors'') come from human geometry: at inference we project an HMR mesh and keep only reliable visible vertices. We then exclude unstable anatomical regions using a precomputed mask, form a dense anchor map by taking per-pixel nearest-\(Z\), and apply a light depth-gradient filter to suppress boundary/occlusion outliers.

The predictor is a UNet encoder--decoder with a ViT bottleneck. The encoder stacks Conv--GroupNorm--GELU blocks with downsampling; the deepest feature is patch-embedded and fed to a Transformer encoder (multi-head self-attention + MLP, norm-first), then reshaped and fused in the decoder via skip connections. Inputs are concatenated maps: the working-domain depth \(x_{\text{proc}}\) (depth or disparity), the anchor map, the anchor mask, and optional RGB. The head outputs two channels \(\tilde{s},\tilde{b}\), constrained as
\(s=1+\alpha_s\tanh(\tilde{s})\) (small scale around 1) and
\(b=\alpha_b\,\mathrm{median}(|x_{\text{proc}}|)\tanh(\tilde{b})\) (scene-adaptive bias).
To stabilize training, \(x_{\text{proc}}\) and anchors are normalized per batch by masked mean-absolute values and clipped.

We use a depth reconstruction loss on valid GT pixels, an anchor consistency loss where anchors exist, and two regularizers on \(s,b\): total variation (spatial smoothness, optionally edge-aware) and spatial variance (keeps fields near global). The total loss is a weighted sum of these terms. Operating in disparity is supported (we convert back to depth for output); tight constraints on \(s\) and regularization on \(s,b\) preserve near-planarity and prevent over-correction. For training, we split a human-centric RGB-D dataset (\textsc{PROX}) into train/test with a 9:1 ratio.

We would like to clarify that our design is motivated by the observation that human reconstruction from MetricHMR is relatively reliable in metric scale and can therefore serve as a geometric anchor for refining monocular depth estimation. Specifically, we use the reconstructed metric human mesh to guide the refinement of MapAnything depth predictions. Unlike prior approaches that apply a simple global scaling factor, we introduce a per-pixel refinement module to correct spatially varying depth errors. This design accounts for the non-uniform inaccuracies commonly observed in monocular depth estimation.

As shown in \cref{tab:depth_prox}, this human-guided refinement improves the metric accuracy of scene depth on the PROX dataset. The current accuracy of monocular depth estimation methods, such as MapAnything, remains insufficient to be reliable. For example, in \cref{fig:depth_refine}, the estimated subject height is 1.59 m while the ground-truth height is 1.72 m. Due to this limitation, we use the predicted depth only as an initialization for scene reconstruction rather than as a constraint to further refine the human pose.




\section{Additional Results}
\subsection{SynFocal dataset.}
Across images with varying camera parameters, MetricHMSR consistently achieves accurate metric distance perception. Recognizing that existing datasets predominantly feature fixed focal lengths, we construct SynFocal, a synthetic dataset of human images that vary in focal length, to further validate the performance of MetricHMSR.

\begin{figure}[htbp]
  \centering
   \includegraphics[width=\linewidth]{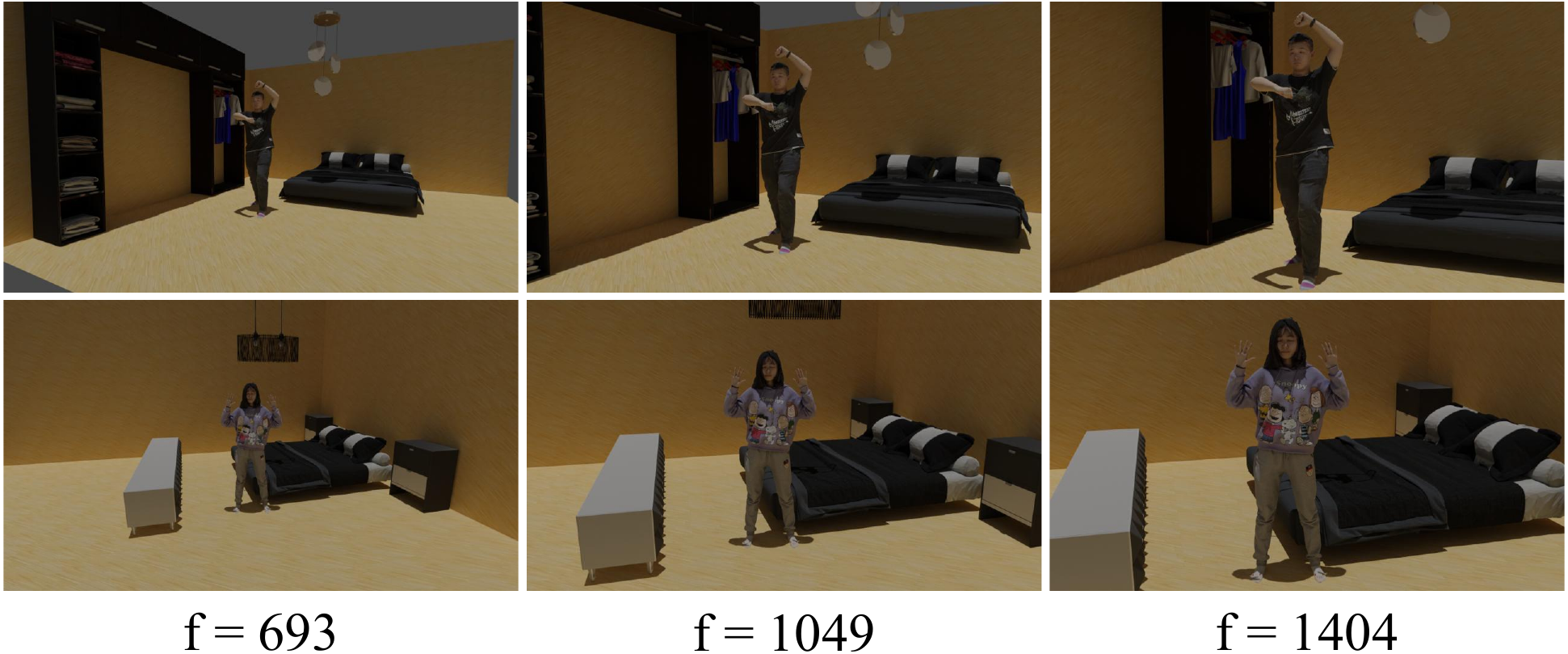}
   \caption{Examples of SynFocal. Each column of images was rendered using a distinct focal length (pixels).
   }
   \label{fig:synfocal}
\end{figure}

Our scenes are constructed using 3D-FRONT~\cite{fu20213dfront}, a synthetic dataset characterized by extensive layout variations and high-fidelity furniture models. Human models are sourced from THuman~\cite{yu2021function4d}, which comprises a large collection of high-quality 3D human models with corresponding SMPL-X~\cite{pavlakos2019smplx} annotations. We used Blender~\cite{blender} to arrange the human and scene in a plausible configuration and rendered images at different focal lengths. \cref{fig:synfocal} shows some examples of SynFocal.

Note that, to the best of our knowledge, there were no human datasets with varying focal lengths before our work. 
Meanwhile, our concurrent work Bedlam2.0~\cite{tesch2025bedlam2} provides richer camera motion and focal length diversity, which can serve as a more comprehensive alternative to SynFocal.

\subsection{Human mesh recovery under varying intrinsics}
For quantitative comparisons, we evaluate performance using two metrics: Root Distance Error (RDE) and Root Position Standard Deviation (RPSD). RDE measures accuracy in the camera coordinate system as the Euclidean distance between predicted and ground-truth root positions, normalized by the distance from the ground-truth root to the origin. RPSD measures robustness as the standard deviation of the Euclidean distances from predicted root joints to their centroid.
We conduct a systematic comparison between MetricHMSR and Human3R~\cite{chen2025human3r} under varying camera intrinsics. 
Quantitative results are reported in \cref{tab:metric_focal}. 
The Results demonstrate that MetricHMSR exhibits stronger robustness to focal length variations than Human3R.
\begin{table}[htbp]
  \centering
    \begin{tabular}{lrr}
    \toprule
      Method     & \multicolumn{1}{l}{RDE} & \multicolumn{1}{l}{RPSD} \\
    \midrule
    Human3R~\cite{chen2025human3r} & 0.14 & 21.6 \\
    MetricHMSR & \textbf{0.10} & \textbf{1.1}\\
    \bottomrule
    \end{tabular}%
    \caption{Quantitative comparisons of 3D position estimation under different intrinsics. RDE is in \% and RPSD is in $\mathrm{cm}$.}
  \label{tab:metric_focal}%
\end{table}%
\subsection{Pseudo-GT Depth for 2D Image Datasets}
\begin{figure}
  \centering
   \includegraphics[width=\linewidth]{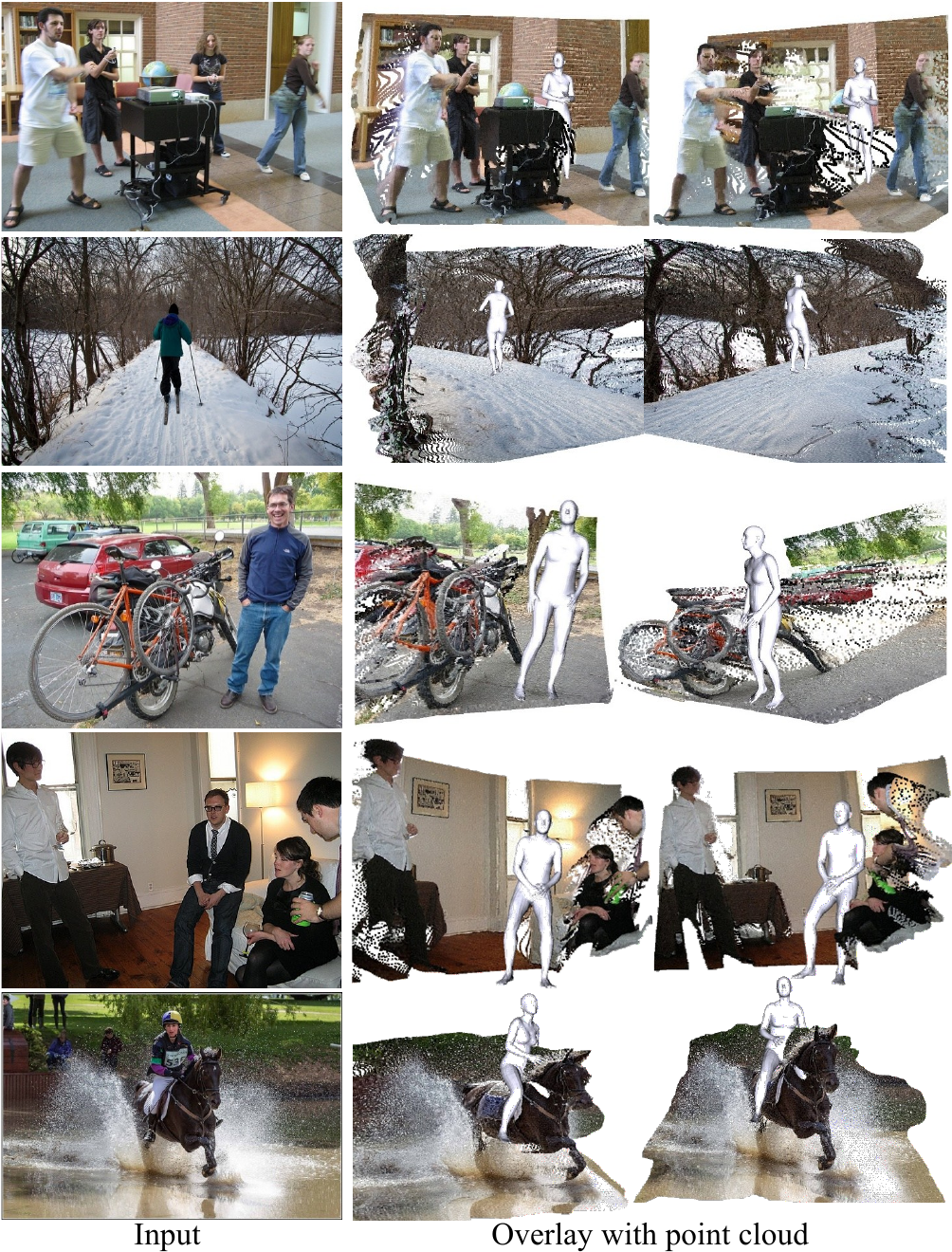}
   \caption{Metric human and scene reconstruction results of in-the-wild data from COCO.}
   \label{fig:5_3Doverlay}
\end{figure}

Based on our Human-Guided Metric Depth Refinement module, we introduce in the main text a new pseudo-GT dataset that supplements background depth information for the commonly used 2D image datasets AI Challenger, COCO, and MPII. 
\cref{fig:5_3Doverlay} illustrates some examples on in-the-wild data from the COCO dataset. We use HumanFoV~\cite{patel2024camerahmr} to estimate the intrinsics of the image, then reconstruct the metric human mesh along with the scene.

More broadly, our method reconstructs metrically accurate humans and scenes from a single image without requiring video, making it possible to leverage the vast amount of image data available across the internet.

\subsection{Metric Comparison}
The metric properties mainly concern metric 3D position and metric body shape. \cref{tab:shape_evaluation} reports the error and stability of the shape, showing that our method outperforms SOTA.
\begin{table}[t]
  \centering
  \setlength{\tabcolsep}{3pt}
  \begin{tabular}{@{}lccccc@{}}
    \toprule
    \multirow{2}{*}{Method} 
      & \multirow{2}{*}{H-MAE} 
      & \multirow{2}{*}{H-MAPE} 
      & \multirow{2}{*}{PVE-T} 
      & \multirow{2}{*}{\shortstack[c]{H\\[-0.5pt]Var.}}
      & \multirow{2}{*}{\shortstack[c]{PVE-T\\[-0.5pt]Var.}} \\
    & & & & & \\
    \midrule
    PromptHMR  & 76.0 & 4.3 & 32.4 & 11.2 & 67.7 \\
    MetricHMSR & \textbf{70.1} & \textbf{3.9} & \textbf{29.9} & \textbf{4.1} & \textbf{27.9} \\
    \bottomrule
  \end{tabular}
  \caption{Metrics on 3DPW. Var. denotes variance.}
  \label{tab:shape_evaluation}
\end{table}

\subsection{Impact of Bbox Scale}
\begin{figure}
  \centering
  \includegraphics[width=\linewidth]{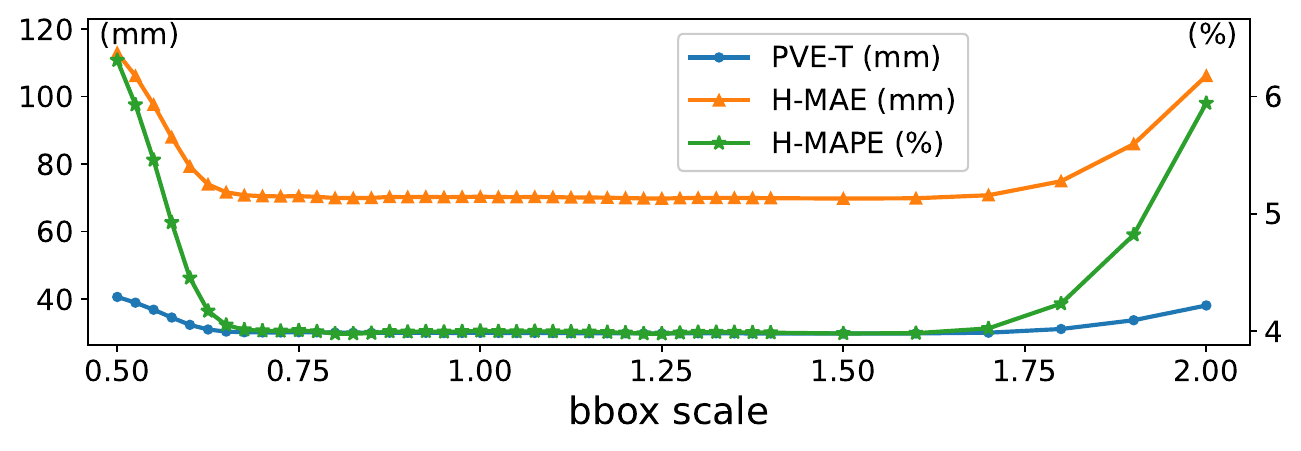}
  \caption{Impact of bounding box scale variation on shape and height accuracy. H: height, T: T-pose, MAE: Mean Absolute Error, MAPE: Mean Absolute Percentage Error.}
  \label{fig:bbox_scale}
\end{figure}

\cref{fig:bbox_scale} plots the variation of the metric against the bbox, where a scale of 1.0 denotes the standard bbox. It is evident that the proposed method demonstrates strong robustness to variations in the bbox.

\begin{figure}[t]
  \centering
   \includegraphics[width=\linewidth]{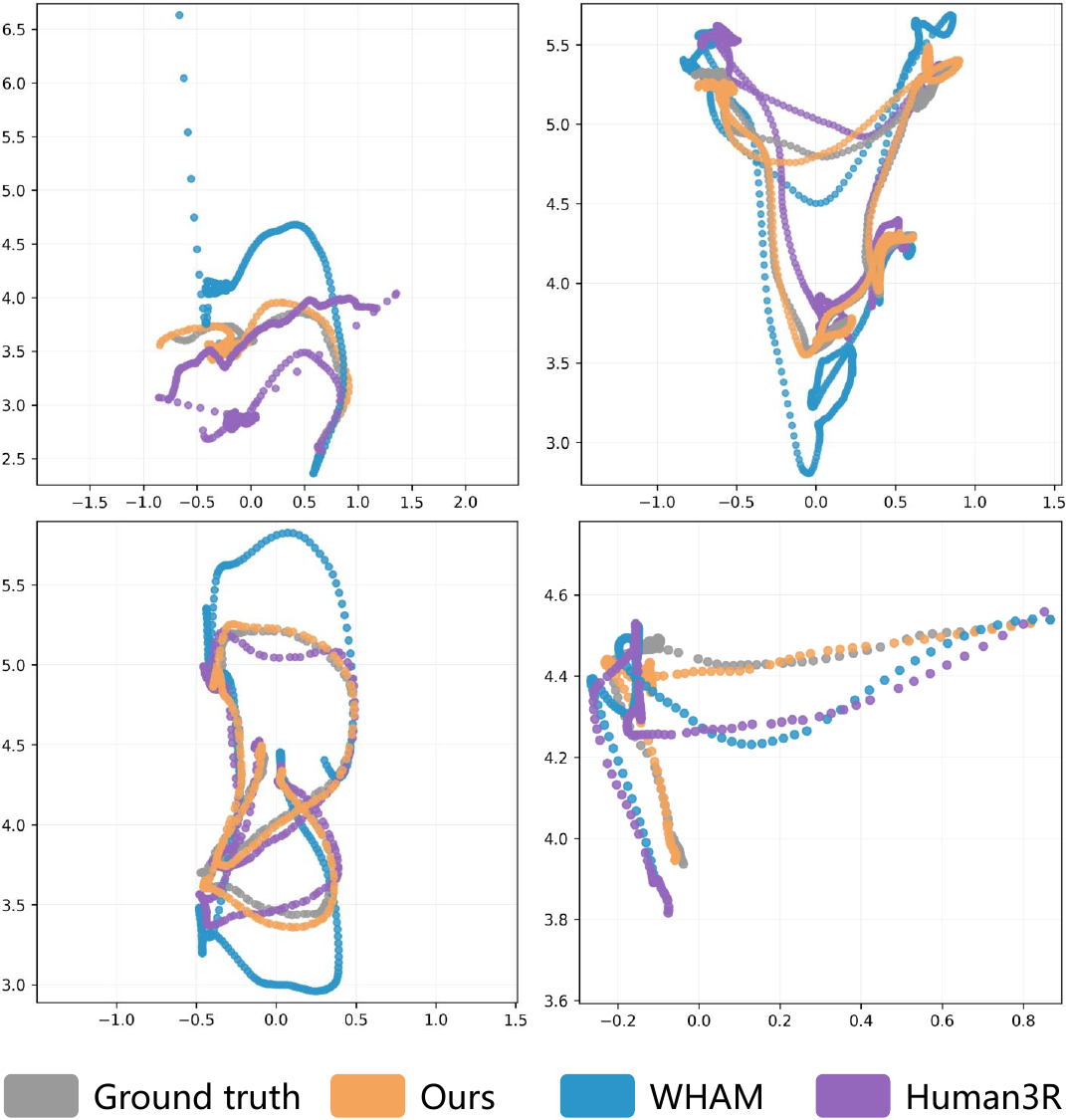}
   \caption{Qualitative comparison of global motion trajectories.}
   \label{fig:Traj}
\end{figure}
\subsection{Trajectory Comparison}
In the quantitative comparisons presented in the main text, our method significantly outperforms existing online approaches in terms of trajectory accuracy, and achieves competitive performance with video-based methods while operating on frame-by-frame inputs rather than video sequences. 

\cref{fig:Traj} shows qualitative comparisons with existing online method on RICH. Our method demonstrates stronger scale awareness and reduced positional drift.

\end{document}